\documentclass[preprint,12pt]{elsarticle}
\usepackage{amssymb}
\usepackage{graphicx}
\usepackage{amsmath}
\usepackage{longtable}
\usepackage{a4wide}
\usepackage{longtable}
\usepackage{algorithm} 
\usepackage{algpseudocode} 
\usepackage{algorithm}
\usepackage{algpseudocode}
\usepackage{mathrsfs}
\usepackage{subcaption}
\usepackage{mathtools}
\usepackage{pifont}
\usepackage{color}
\usepackage{lineno}
\usepackage{makecell} 
\usepackage{pdflscape}
\usepackage{adjustbox}
\usepackage[utf8]{inputenc}
\usepackage{tabularx}
\usepackage{blindtext}
\usepackage{setspace}
\usepackage{multirow}
\usepackage{notoccite} 
\usepackage{lscape} 
\usepackage{caption} 
\usepackage{mwe}
\usepackage{booktabs}
\usepackage{hyperref}
\usepackage{amsthm}

\newcommand{\cmark}{\ding{51}}%
\newcommand{\xmark}{\ding{55}}%
\newcommand{\RNum}[1]{\lowercase\expandafter{\romannumeral #1\relax}}
\newcommand{\RNumU}[1]{\uppercase\expandafter{\romannumeral #1\relax}}
\usepackage{natbib}
\journal{Elsevier}

\begin{document}
\begin{frontmatter}
\title{GRVFL-MV: Graph Random Vector Functional Link Based on Multi-View Learning}
\author[inst1]{M. Tanveer\texorpdfstring{\corref}{corref}{Correspondingauthor}}\ead{mtanveer@iiti.ac.in}
\author[inst1]{R. K. Sharma}\ead{msc2203141001@iiti.ac.in}
\author[inst1]{M. Sajid }\ead{phd2101241003@iiti.ac.in}
\author[inst1]{A. Quadir}\ead{mscphd2207141002@iiti.ac.in}

\affiliation[inst1]{organization={Department of Mathematics, Indian Institute of Technology Indore},
            addressline={Simrol}, 
            city={Indore},
            postcode={453552}, 
            state={Madhya Pradesh},
            country={India}}
            \cortext[Correspondingauthor]{Corresponding author}
\begin{abstract}
The classification performance of the random vector functional link (RVFL), a randomized neural network, has been widely acknowledged. However, due to its shallow learning nature, RVFL often fails to consider all the relevant information available in a dataset. Additionally, it overlooks the geometrical properties of the dataset. To address these limitations, a novel graph random vector functional link based on multi-view learning (GRVFL-MV) model is proposed. The proposed model is trained on multiple views, incorporating the concept of multiview learning (MVL), and it also incorporates the geometrical properties of all the views using the graph embedding (GE) framework. The fusion of RVFL networks, MVL, and GE framework enables our proposed model to achieve the following: i) \textit{efficient learning}: by leveraging the topology of RVFL, our proposed model can efficiently capture nonlinear relationships within the multi-view data, facilitating efficient and accurate predictions; ii) \textit{comprehensive representation}: fusing information from diverse perspectives enhance the proposed model's ability to capture complex patterns and relationships within the data, thereby improving the model's overall generalization performance; and iii) \textit{structural awareness}: by employing the GE framework, our proposed model leverages the original data distribution of the dataset by naturally exploiting both intrinsic and penalty subspace learning criteria. The evaluation of the proposed GRVFL-MV model on various datasets, including 29 UCI and KEEL datasets, 50 datasets from Corel5k, and 45 datasets from AwA, demonstrates its superior performance compared to baseline models. These results highlight the enhanced generalization capabilities of the proposed GRVFL-MV model across a diverse range of datasets.

\end{abstract}

\end{frontmatter}
\section{Introduction}
The artificial neural networks (ANNs) belong to the class of non-parametric learning methods that are used for estimating or approximating functions that may depend on a large number of inputs and outputs \cite{zhang2016survey}. Inspired by the topology of biological neural networks, ANNs have found applications in diverse domains, such as image recognition \cite{shafiq2022deep}, Alzheimer's disease diagnosis \cite{tanveer2024fuzzy}, stock price prediction \cite{gulmez2023stock}, and so on. The efficacy of ANNs depends upon several factors, including the quality and quantity of data, computational resources, and the effectiveness of underlying algorithms \cite{cao2018review}. The  ANNs are trained by optimizing a cost function, which quantifies the disparity between model predictions and actual observations. The adjustment of the network's weights and biases is facilitated through back-propagation, employing an iterative technique known as the gradient descent (GD) method. To enhance the agreement between predicted outcomes and actual observations, the optimization process requires fine-tuning the network's weights and biases. Nevertheless, GD-based algorithms come with inherent limitations, including slow convergence \cite{jacobs1988increased}, difficulty in achieving global minima \cite{gori1992problem} and heightened sensitivity to the selection of learning rate and the point of initialization. 

Randomized neural networks (RNNs) \cite{201708} have emerged as a unique solution to overcome the difficulties faced in training ANNs using gradient descent (GD) methods. Unlike conventional ANNs, RNNs adopt a distinct approach where the network's weights and biases are randomly selected from a specified range and remain constant throughout the training process. The determination of output layer parameters in RNNs is achieved through a closed-form solution \cite{suganthan2021origins}, as opposed to iterative-based optimization techniques commonly employed in traditional ANNs.
The prominent examples of RNNs include the random vector functional link (RVFL) network \cite{pao1994learning} and the RVFL without direct link (RVFLwoDL) (also known as extreme learning machine (ELM))\cite{huang2006extreme}. RVFL has garnered significant attention within the realm of RNNs due to its distinctive features, including direct connections between input and output layers that improve generalization performance, a straightforward architecture aligned with Occam's Razor principle and PAC learning theory  \cite{kearns1994introduction}, a reduced number of parameters, and the universal approximation capability \cite{igelnik1995stochastic}. Furthermore, the direct connections help to regulate the randomness in RVFL \cite{zhang2016survey}, while the output parameters are computed analytically through methods like pseudo-inverse or least-square techniques.

Various enhancements have been implemented to optimize the learning capabilities of RVFL, with notable advancements such as the sparse pre-trained random vector functional link (SP-RVFL) network \cite{zhang2019unsupervised} and the discriminative manifold RVFL neural network (DM-RVFLNN) \cite{li2021discriminative}. The SP-RVFL network utilizes diverse weight initialization methods to improve the generalization performance of standard RVFL models, while the DM-RVFLNN integrates manifold learning with a soft label matrix to enhance the model's discriminative capacity by increasing the margin between samples from different classes. In addition, techniques such as manifold learning based on an in-class similarity graph have been employed to enhance the compactness and similarity of samples within the same class. To address noise and outliers, authors in \cite{cui2017received} proposed an RVFL-based approach that incorporates a novel feature selection (FS) technique, enhancing the efficiency and robustness of RVFL through the augmented Lagrangian method. Furthermore, similar to support vector machines and their variants \cite{10759815}, incorporating a kernel function into RVFL has resulted in the kernel exponentially extended RVFL network (KERVFLN) \cite{chakravorti2020non}, demonstrating the adaptability and potential of RVFL in various machine learning applications. Recent advancements in RVFL theory include RVFL+, kernel RVFL+ (KRVFL+) \cite{zhang2020new}, and incremental learning paradigm with privileged information for RVFL (IRVFL+), showcasing the ongoing evolution and versatility of RVFL models in addressing diverse machine-learning challenges, including multilabel classification tasks using RNNs. Some recent advancements in RVFL theory include ensemble deep variant of RVFL based on fuzzy inference system \cite{10552388}, kernel ridge regression-based randomized network for brain age classification and estimation \cite{10405861}, domain-incremental learning without forgetting based on RVFL \cite{liu2024domain},  etc. 

In various real-world applications, a multitude of characteristics is often observed, requiring representation through multiple feature sets. This leads to the prevalence of multiview data, where information from various measurement methods is collected to comprehensively capture the nuances of each example rather than relying solely on a single feature set. For instance, a picture can be described by color or texture features, and a person can be identified by face or fingerprints \cite{xu2017re}. Web pages serve as a quintessential example of multimodal data, where one feature vector encapsulates the words within the webpage text, while another feature vector captures the words present in the links pointing to the webpage from other pages. While individual views may suffice for specific learning tasks, there is potential for enhancement by fusing insights from multiple data representations \cite{houthuys2018multi}. Multiview learning (MVL), a well-established collection of techniques, holds significant potential as multi-modal datasets become increasingly accessible \cite{xu2022multi}. MVL models are often developed under the supervision of the consensus or complimentary principles \cite{quadir2024enhancing} to ensure the effectiveness of an algorithm.  The consensus principle aims to enhance the performance of classifiers for each view by maximizing consistency across multiple viewpoints. Conversely, the complementarity principle emphasizes the importance of providing complementary data from diverse perspectives to offer a comprehensive and accurate description of the object.

The incorporation of the graph embedding (GE) framework \cite{yan2006graph} into the RVFL model enhances its learning capabilities by embedding the intrinsic geometric structure of the data into the feature space. This integration enables the RVFL model to preserve local and global geometrical relationships, which were previously neglected, thereby improving its representational power and generalization performance.To incorporate the geometrical relationships within a dataset, authors in \cite{malik2022graph} introduced the graph-embedded intuitionistic fuzzy weighted RVFL (GE-IFWRVFL) model. This model not only integrates the intuitionistic fuzzy (IF) membership scheme to handle noisy data and outliers but also preserves the geometric characteristics through the GE framework. The fusion of IF and GE with RVFL has proven to enhance the resilience and performance of the RVFL significantly. 
The experimental results from \cite{10431593}, along with studies in \cite{MALIK2023110377}, strongly support the notion that the performance of RVFL is greatly improved through the fusion of the GE framework. This emphasizes the importance of considering the geometric attributes of a dataset in the learning process to enhance the overall performance of RVFL.

RVFL is categorized as a shallow learning algorithm due to its single hidden layer, which may potentially hinder its ability to fully grasp complex patterns and subtle nuances in a dataset. While RVFL may excel in specific classification tasks, its architecture could limit its capability to accurately interpret intricate patterns or extract detailed features from the dataset. The simplicity of RVFL's architecture offers advantages in terms of computational efficiency and ease of implementation. However, when confronted with datasets containing intricate patterns or nuanced relationships among variables, more sophisticated learning frameworks are necessary to effectively capture the underlying complexities inherent in the dataset. To address the constraints associated with RVFL’s shallow architecture, adopting a multiview learning (MVL) framework presents a viable solution. By training RVFL on multiple views of the data, the model can integrate diverse perspectives, thereby mitigating its structural limitations and enhancing its ability to discern intricate patterns. Leveraging MVL not only enriches the representational capacity of RVFL but also significantly improves its learning efficiency, making it more suitable for complex data analysis tasks.

Recognizing the importance of a dataset's inherent geometrical characteristics and MVL, we propose the novel graph random vector functional link based on multi-view learning (GRVFL-MV). The GRVFL-MV method incorporates the intrinsic and penalty graphical representations of multiview datasets within the GE framework, thereby capturing the geometrical properties of the multiview data. By fusing the MVL and GE framework with RVFL, the learning capabilities of RVFL are greatly enhanced, allowing it to effectively tackle classification challenges. In order to ensure simplicity and effectiveness, the proposed GRVFL-MV model incorporates two views at a time. This strategy allows for a balance between complexity and performance, as it utilizes the complementary information from these two views to enhance effectiveness while still maintaining simplicity. By leveraging information from multiple views, GRVFL-MV significantly improves the classification performance of RVFL by integrating both geometric and discriminative information from both views using intrinsic and penalty-based subspace learning criteria within the GE framework. 

The main contributions of this research are highlighted below:
\begin{itemize}

    \item We present a generic framework that fuses random vector functional link (RVFL) with multiview learning (MVL) and graph embedding (GE) framework. This novel model is called the graph random vector functional neural network based on multi-view learning (GRVFL-MV).

    \item The proposed GRVFL-MV model is developed upon the foundation of the RVFL architecture, which is known for its simplicity and efficiency. However, it goes beyond the limitations of shallow learning of RVFL by integrating the concept of multiview learning (MVL). By leveraging multiple views, the model aims to enhance classification performance using multiple perspectives of the dataset.

    \item The proposed GRVFL-MV model integrates geometrical information from multiview data by embedding intrinsic and penalty subspace learning (SL) criteria within the GE framework. It employs local Fisher discriminant analysis (LFDA) \cite{sugiyama2007dimensionality} and graph regularization parameters to effectively utilize the GE framework.

    \item To trade off the information from multiple views, our proposed mathematical formulation includes a coupling term. This coupling term helps to mitigate errors between the views, resulting in improved generalization performance.

    \item Our experiments encompassed 29 UCI and KEEL datasets, 50 datasets from Corel5k, and 45 datasets from AwA. Through comprehensive statistical analyses, we demonstrate the superior generalization performance of our proposed model when compared to baseline models.
\end{itemize}
The structure of the paper is as follows: Section \ref{sec-2} offers a detailed overview of the preliminary works that form the foundation of this study. Section \ref{sec-3} delves into the mathematical formulation and computational complexity of the proposed model. Section \ref{sec-5} discusses the experimental results alongside statistical analyses. Section \ref{comprehensive analysis} provides an in-depth evaluation of the proposed model's performance across different dataset types. Finally, Section \ref{sec-6} concludes the paper with key findings and suggestions for future research directions.

\section{Preliminary Works}
\label{sec-2}
In this section, we start with fixing some notations and then provide a detailed description of random vector functional (RVFL) \cite{pao1994learning}. We also give the literature overview of multiview learning (MVL) and review the graph embedding \cite{xu2021understanding} framework. 

\subsection{Notations}
 For simplicity, we consider the notations w.r.t. two-views. Let the two-view training dataset $ S = \bigl\{(x_i^A, x_i^B,y_i)| x_i^A \in \mathbb{R}^{1 \times n}, x_i^B \in \mathbb{R}^{1 \times m},\, y_i \in \{-1,1\}, i = 1, 2, \cdots, l\bigl\}$, where $x_i^A$  and $x_i^B$ represents the feature vector of each data sample corresponding to $view-A$ and $view-B$, respectively and $y_i$ represents the label of the individual data for both the views. Let $X_A \in \mathbb{R}^{l \times n} $ and $X_B \in \mathbb{R}^{l \times m}$ be the input matrices for $view-A$ and $view-B$, respectively and $ Y_{true} \in \mathbb{R}^{l \times 2}$ represents the one hot encoding matrix of the output. $(\cdot)^{t}$ represents the transpose operator. 
 
 \subsection{Random Vector Functional Link Neural Network (RVFL)}
The RVFL \cite{pao1994learning} model consists of an input layer, a hidden layer, and an output layer. The input and output layers are connected through the hidden layer, which acts as a bridge between them. Notably, the original features are also directly passed to the output layer because of the direct connections between the input and output layers. During the training process, the weights connecting the input and hidden layers, as well as the biases at the hidden layer, are randomly generated. Once generated, these parameters remain fixed and do not require any adjustments during the training phase. In terms of determining the output weight matrix connecting the input layer and the hidden layer to the output layer, the least squares or pseudo-inverse methods are employed, providing an analytical solution. \\
\begin{figure}[!ht]
       \centering
        \includegraphics[width=9cm, height=6cm]{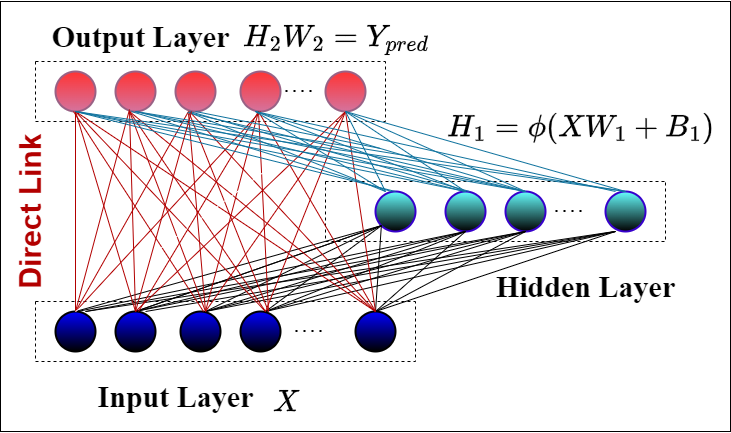}
        \caption{The architecture of RVFL model.}
        \label{fig:enter-label}
\end{figure}
Consider a training matrix $X$ with dimensions $l \times p$. Let $W_1$ be a weight matrix with dimensions $p \times h$, where the values are randomly generated from a uniform distribution within the range of $[-1, 1]$ and  $B_1 \in \mathbb{R}^{l \times h}$ (all the columns are identical) be the randomly generated bias matrix. To obtain the hidden layer matrix, also called the randomized feature layer,  denoted as $H_1$, we apply a nonlinear activation function $\phi$ to the matrix $XW_1 + B_1$.  Consequently, the hidden layer matrix $H_1$ can be expressed as:
\begin{equation}
    H_1 = \phi(XW_1 + B_1).
\end{equation}

Therefore, the matrix $H_1$ can be represented as:

$$
H_1 = \begin{bmatrix}
    \phi(x_1w_1 + b_1) &  \cdots &  \phi(x_1w_h + b_h) \\
    \vdots  &   \vdots & \vdots \\ 
    \phi(x_pw_1 + b_1) &  \cdots &  \phi(x_pw_h + b_h)
\end{bmatrix},
$$

where $x_i \in \mathbb{R}^{1 \times p}$ and $w_j \in \mathbb{R}^{p \times 1}$ represent the $i^{th}$ row of $X$ and the $j^{th}$ column of $W_1$, respectively. The term $b_j$ represents the bias of the $j^{th}$ hidden node. 

Let $H_2$ be a concatenated matrix of features from the input and hidden layers, represented as $H_2 = \begin{bmatrix} X & H_1 \end{bmatrix} \in \mathbb{R}^{l \times (p+h)}$. Here, $X$ represents the input features, and $H_1$ represents the output of the hidden layer. Let $W_2$ denote the weights matrix connecting the concatenation of input ($X$) and hidden ($H_1$) layers to the output layer, with dimensions $W_2 \in \mathbb{R}^{(p+h) \times 2}$. The predicted output matrix $Y_{pred} \in \mathbb{R}^{l \times 2}$ of the RVFL is calculated using the following matrix equation:
\begin{equation}
    \label{eqn:2}
H_2W_2 = Y_{pred}.
\end{equation}

The optimization problem of (\ref{eqn:2}) is given as:
\begin{align}
    \label{eqn:3}
    \underset{W_2}{\text{min}}\hspace{0.2cm}\frac{1}{2}||W_2||_2^2 + \frac{c}{2}||\xi||_2^2 \nonumber \\
    \text {s.t.} \hspace{0.2cm} H_2W_2 - Y_{true} =\xi,
\end{align}
where $c$ is a regularization parameter.

The solution of (\ref{eqn:3}) is given by:
\begin{equation}
   (W_2)_{min} =   \begin{cases} 
      ({H_2}^{t}H_2 + \frac{1}{c}I)^{-1}{H_2}^{t}Y_{true} , & (p+h)\leq l \\
      {H_2}^{t}({H_2}H_2^{t} + \frac{1}{c}I)^{-1}Y_{true} , & l < (p+h) 
      \end{cases}
\end{equation}

\subsection{Graph Embedding (GE)}
The concept of graph embedding (GE) framework \cite{xu2021understanding} aims to capture the underlying graphical structure of data in a vector space. In this framework, a given input matrix  $X$  is utilized to define the intrinsic graph \(\mathcal{G}^{int} = \{X, \Delta^{int}\}\) and the penalty graph \(\mathcal{G}^{pen} = \{X, \Delta^{pen}\}\) for the purpose of subspace learning (SL) \cite{yan2006graph}. The similarity weight matrix \(\Delta^{int} \in \mathbb{R}^{l \times l}\) incorporates weights that represent the pairwise relationships between the vertices in \( X \). Conversely, the penalty weight matrix \(\Delta^{pen} \in \mathbb{R}^{l \times l}\) assigns penalties to specific relationships among the graph vertices. The optimization problem for graph embedding is formulated as follows:
\begin{align}{}
\label{eqn:4}
{v^*} &= \underset{tr({v_0}^{t}X^{t}\mathbb{U}Xv_0) = d}{\text{argmin}}\sum_{k \neq l} || {v_0}^{t}x_k -  {v_0}^{t}x_l ||\Delta_{kl}^{int} \nonumber \\
    & = \underset{tr({v_0}^{t}X^{t}\mathbb{U}Xv_0) = d}{\text{argmin}}tr({v_0}^{t}X^{t}\mathbb{L}Xv_0)
\end{align}
where $tr(\cdot)$ is the trace operator and $d$ is a constant value. \( \mathbb{L} = \mathbb{D} - \Delta^{{int}} \in \mathbb{R}^{l\times l} \) is a representation of the Laplacian matrix of the intrinsic graph \( \mathcal{G}^{\text{int}} \), with the diagonal elements of \( \mathbb{D} \) being defined as \( \mathbb{D}_{kk} = \sum_l \Delta^{int}_{kl} \).  Moreover, \( \mathbb{U} = \mathbb{L}^p = \mathbb{D}^p - \Delta^{pen} \) serves as the Laplacian matrix of the penalty graph \( \mathcal{G}^{\text{pen}} \).  The matrix \( v_0 \) is associated with the projection matrix. Equation (\ref{eqn:4}) can be simplified to a generalized eigenvalue problem as shown in the form:
\begin{equation}
    G_{int}s = {\lambda}G_{pen}s , 
\end{equation}
where \( G_{int} = X^{t}\mathbb{L}X \) and \( G_{pen} = X^{t}\mathbb{U}X \). This simplification implies that the transformation matrix will be constructed from the eigenvectors of the matrix \( G = G_{pen}^{-1} G_{int} \), where \( G \) integrates the intrinsic and penalty graph connections of the data samples.
\subsection{Literature Review on Multiview Learning (MVL)}
Multiview learning (MVL) is an emerging research paradigm with substantial potential to enhance generalization performance across diverse learning tasks. By integrating multiple feature representations, each capturing distinct and complementary information, MVL effectively enriches the learned feature space, leading to improved model robustness and predictive accuracy \cite{tang2020cgd}. The abundance of diverse data types in practical applications has led to the development of MVL. In real-world scenarios, samples from different perspectives may reside in distinct spaces or exhibit vastly different distributions, often due to significant disparities between views \cite{li2016low}. However, conventional approaches typically address such data by employing a cascade strategy, wherein multiview data is amalgamated into a single-view format through the concatenation of heterogeneous feature spaces into a homogeneous one. However, this cascading approach overlooks the distinctive statistical properties of each view and is plagued by the curse of dimensionality problems. One notable advantage of MVL is that it can enhance the performance of a standard single-view approach by leveraging manually generated multiple views.

In MVL, a distinct function is developed for each view, with the overall goal of constructing a unified function to optimize all individual functions jointly. This approach enhances the generalization performance of the framework across multiple views. According to \cite{zhao2017multi}, MVL models can be categorized into three main groups: co-training style algorithms, co-regularization style algorithms, and margin consistency style algorithms. Co-training style algorithms focus on enhancing mutual agreement among diverse views. Conversely, co-regularization style algorithms aim to minimize discrepancies during the learning process. MVL learning has been successfully implemented in various hyper-plane based classifiers, such as SVM-2K \cite{farquhar2005two}, multiview twin support vector machine \cite{xie2015multi} (MvTSVM), deep multi-view multiclass TSVM \cite{xie2023deep}, etc. A notable application of MVL has been found in predicting DNA-binding protein using RVFL \cite{quadir2024multiview}. Furthermore, MVL has demonstrated its effectiveness across various application scenarios, including enhancing image classification, annotation, and retrieval performance \cite{chen2010predictive}, predicting financial distress \cite{sun2021multi}, and identifying product adoption intentions from social media data \cite{zhang2021predicting}. MVL is now widely used in many various domains \cite{YAO2017236} and research endeavors.

\section{The Proposed Graph Random Vector Functional Link Neural Network based on Multi-View Learning (GRVFL-MV)}
\label{sec-3}
In this section, we introduce a novel GRVFL-MV, which fuses the RVFL with MVL. Our proposed model is structured to leverage the strengths of RVFL while incorporating both intrinsic and penalty graphical representations of multiview data through the GE framework. To derive the network output weights, the optimization process integrates subspace learning (SL) criteria, incorporating both intrinsic and penalty-based SL methodologies within the GE framework. To preserve the geometric structure effectively, we incorporate local Fisher discriminant analysis (LFDA) \cite{sugiyama2007dimensionality} into the GE framework. Additionally, we introduce a regularization parameter for GE to enhance the learning process further. To trade off the error between multiple views, we introduce a coupling term in the primal optimization problem of the proposed model. This term minimizes the combined error from both views, resulting in an improved generalization performance for our model. The proposed optimization problem of the GRVFL-MV model is presented below:
\begin{align}{}
\label{eq:1}
\underset{\beta_1, \beta_2}{\text{min}}\hspace{0.2cm} &\frac{c_1}{2}||\xi_1||_2^2 + \frac{c_2}{2}||\xi_2||_2^2 + \frac{c_3}{2} ||\beta_1||_2^2+  \frac{1}{2}||\beta_2||_2^2 +  \frac{\theta_1}{2}||G_1^{1/2}\beta_1||_2^2 +  \frac{\theta_2}{2}||G_2^{1/2}\beta_2||_2^2 + \rho{\xi_1^{t}}\xi_2 \nonumber  \\
&\text {s.t.} \hspace{0.2cm} Z_1\beta_1 - Y_{true}=\xi_1 \hspace{0.1cm} \text{and} \hspace{0.1cm} Z_2\beta_2 - Y_{true}  =\xi_2,\\ \nonumber \\
\label{eq8}
&\text{where}\hspace{0.1cm} Z_1= \begin{bmatrix}X_A & H_A \end{bmatrix} \hspace{0.1cm} \text{and}\hspace{0.1cm} Z_2 =\begin{bmatrix}X_B & H_B \end{bmatrix}. \end{align}$H_A = \phi(X_AW_A + B_A)$ and $H_B = \phi(X_BW_B + B_B)$, where $\phi(\cdot)$ is a non-linear activation function. $W_A$ and $B_A$ are the randomly initialized weight and bias matrices for $view-A$ and $W_B$ and $B_B$ for $view-B$, respectively. \\
Optimization problem (\ref{eq:1}) has the following components:
\begin{enumerate}
    \item Connections between the concatenated matrix $Z_1$ and the output layer are established by the weight matrix \( \beta_1 \), while the weight matrix \( \beta_2 \) establishes connections between the $Z_2$ and the output layer. Minimizing $||\beta_1||_2^2$ and $||\beta_2||_2^2$ incorporates the structural risk minimization (SRM) principle.  

    \item \( G_1 \) and \( G_2 \) represent the graph embedding matrices w.r.t. $view-A$ and $view-B$, respectively, and minimizing $||G_1^{1/2}\beta_1||_2^2$ and $||G_2^{1/2}\beta_2||_2^2$ facilitates the preservation of graph structural relationships between data points. 

    \item\( \xi_1 \) and \( \xi_2 \) represent the empirical errors in $view-A $ and $view-B$, respectively. Minimizing $ ||\xi_1||_2^2 $ and $ ||\xi_2||_2^2 $ leads to the reduction of empirical errors in both views. 

    \item The term $\xi_1^{t}\xi_2$ acts as a coupling term that fuses information from both views and promotes the simultaneous minimization of errors in both views.

    \item  The variables \( \theta_1 \) and \( \theta_2 \) are graph regularization parameters, while \( c_1 \), \( c_2 \), and \( c_3 \) represent the regularization parameters and   \( \rho \) is the coupling parameter.
\end{enumerate}
The lagrangian of (\ref{eq:1}) is:
\begin{align}{}
\label{eq:2}
L = \frac{c_1}{2}||\xi_1||^2 +  &\frac{c_2}{2}||\xi_2||^2 + \frac{c_3}{2}||\beta_1||^2+  \frac{1}{2}||\beta_2||^2 +   \frac{\theta_1}{2}||G_1^{1/2}\beta_1||^2 + \frac{\theta_2}{2}||G_2^{1/2}\beta_2||^2 + \nonumber\\ &\rho{\xi_1^{t}}\xi_2   - {\alpha_1}^{t}(Z_1\beta_1 - Y_{true} -\xi_1)
- {\alpha_2}^{t}(Z_2\beta_2 - Y_{true} -\xi_2).
\end{align}
Partially differentiating w.r.t. $\xi_1, \xi_2, \beta_1, \beta_2, \alpha_1,$ and $\alpha_2$ we get
\begin{align}{}
\label{eq:3}
\frac{\partial L}{\partial \xi_1} & = c_1\xi_1 + \rho\xi_2 + \alpha_1 = 0, \\
\label{eq:4}
\frac{\partial L}{\partial \xi_2} & = c_2\xi_2 + \rho\xi_1 + \alpha_2 = 0, \\
\label{eq:5}
\frac{\partial L}{\partial \beta_1} & = c_3\beta_1 + \theta_1G_1\beta_1 - {Z_1}^{t}\alpha_1= 0, \\
\label{eq:6}
\frac{\partial L}{\partial \beta_2} & = \beta_2 + \theta_2G_2\beta_2 - {Z_2}^{t}\alpha_2= 0, \\
\label{eq:7}
\frac{\partial L}{\partial \alpha_1} &= Z_1\beta_1 - Y_{true} -\xi_1 = 0 ,\\
\label{eq:8}
\frac{\partial L}{\partial \alpha_2} &= Z_2\beta_2 - Y_{true} -\xi_2 = 0.
\end{align}
Substituting (\ref{eq:3}) and (\ref{eq:4}) in (\ref{eq:5}) and (\ref{eq:6}) respectively, we get
\begin{align}
\label{eq:9}
c_3\beta_1 + \theta_1G_1\beta_1 + {Z_1}^{t}(c_1\xi_1 + \rho\xi_2) = 0,\\
\label{eq:10}
\beta_2 + \theta_2G_2\beta_2 + {Z_2}^{t}(c_2\xi_2 + \rho\xi_1) = 0.
\end{align}
Substituting the value of (\ref{eq:7}) and (\ref{eq:8}) in (\ref{eq:9}) and (\ref{eq:10}), we get
\begin{align}
\label{eq:11}
c_3\beta_1 + \theta_1G_1\beta_1 + {Z_1}^{t}\left(c_1(Z_1\beta_1 - Y_{true}) + \rho(Z_2\beta_2 - Y_{true})\right) = 0,\\
\label{eq:12}
\beta_2 + \theta_2G_2\beta_2 + {Z_2}^{t}\left(c_2(Z_1\beta_1 - Y_{true}) +  \rho(Z_2\beta_2 - Y_{true})\right) = 0. 
\end{align}
On simplifying, we get
\begin{align}
\label{eq:13}
c_3\beta_1 + \theta_1G_1\beta_1 + c_1{Z_1}^{t}Z_1\beta_1 +   \rho{Z_1}^{t}Z_2\beta_2    = {Z_1}^{t}(c_1 + \rho)Y_{true},\\
\label{eq:14}
\beta_2 + \theta_2G_2\beta_2 + c_2{Z_2}^{t}Z_2\beta_2 +  \rho{Z_2}^{t}Z_1\beta_1  = {Z_2}^{t}(c_2 + \rho)Y_{true}.
\end{align}
Finally, we get
\begin{align}
\label{eq:15}
(c_3I_1 + \theta_1G_1 + c_1{Z_1}^{t}Z_1)\beta_1 +   (\rho{Z_1}^{t}Z_2)\beta_2  = {Z_1}^{t}(c_1 + \rho)Y_{true},\\
\label{eq:16}
(\rho{Z_2}^{t}Z_1)\beta_1 + (I_2 + \theta_2G_2 + c_2{Z_2}^{t}Z_2)\beta_2  = {Z_2}^{t}(c_2 + \rho)Y_{true},
\end{align}
where $I_1$ and $I_2$ are the identity matrices of the appropriate dimension. Converting into matrix form, we get
\begin{align}
\begin{bmatrix}
    c_3I_1 + \theta_1G_1 + c_1{Z_1}^{t}Z_1 & \rho{Z_1}^{t}Z_2 \\
    \rho{Z_2}^{t}Z_1 & I_2 + \theta_2G_2 + c_2{Z_2}^{t}Z_2
    \end{bmatrix}   \begin{bmatrix}
        \beta_1 \\ \beta_2
    \end{bmatrix}   = 
    \begin{bmatrix}
        {Z_1}^{t}(c_1 + \rho) \\
        {Z_2}^{t}(c_2 + \rho)
    \end{bmatrix}Y_{true}.
\end{align}

\begin{align}
\label{eq.24}
\begin{bmatrix}
        \beta_1 \\ \beta_2
    \end{bmatrix}   =  \begin{bmatrix}
    c_3I_1 + \theta_1G_1 + c_1{Z_1}^{t}Z_1 & \rho{Z_1}^{t}Z_2 \\
    \rho{Z_2}^{t}Z_1 & I_2 + \theta_2G_2 + c_2{Z_2}^{t}Z_2
    \end{bmatrix}^{-1} 
  \begin{bmatrix}
        {Z_1}^{t}(c_1 + \rho) \\
        {Z_2}^{t}(c_2 + \rho)
    \end{bmatrix}Y_{true}.
\end{align}
For a new data point $x$ having representation $x_A$ and $x_B$ w.r.t.  $view-A$ and $view-B$, respectively, we give the classification function as follows:
\begin{align}
\label{eq:19}
    & class(x) = \underset{i \in \{1, 2\}}{\arg\max} \{y_{c_i}\}, \\ 
     & \hspace{-0.4cm} \text{where} \nonumber \\ 
      & y_{c} = \frac{1}{2}\left( [x_A\hspace{0.2cm} \phi(x_AW_A + b_A)]\beta_1  +  [x_B\hspace{0.2cm} \phi(x_BW_B + b_B)]\beta_2)  \right) \hspace{0.1cm}  \nonumber \\
      & \text{and} \hspace{0.1cm} y_c = (y_{c_1}, y_{c_2}).\nonumber
\end{align}
where $W_A$ and $W_B$ are the randomly generated weights matrices, and $b_A$ and $b_B$ represents the bias column vectors of the bias matrices $B_A$ and $B_B$, respectively.  Since, all the columns of the bias matrix $B_A$ are identical, hence $b_A$ can be chosen to be any column of $B_A$. A similar argument follows for $b_B$. The algorithm for the proposed GRVFL-MV is presented in Algorithm 1.
\subsection{LFDA Under the GE Framework}
The intrinsic as well as the penalty graphs are constructed using concatenated matrices \( Z_1 \) and \( Z_2 \). Specifically, for $Z_1$, we have $\mathcal{G}_1^{int} = \lbrace Z_1, \prescript{}{1}\Delta^{int}\rbrace$ and $\mathcal{G}_1^{pen} = \lbrace Z_1, \prescript{}{1}\Delta^{pen}\rbrace$. Similarly for $Z_2$, we have $\mathcal{G}_2^{int} = \lbrace Z_2, \prescript{}{2}\Delta^{int}\rbrace$ and $\mathcal{G}_1^{pen} = \lbrace Z_2, \prescript{}{2}\Delta^{pen}\rbrace$. Hence,  the intrinsic graph  is denoted as \( G_{int}^1 = {Z_1}^{t}\mathbb{L}_1{Z_1} \) and the penalty graph as \( G_{pen}^1 = {Z_1}^{t}\mathbb{U}_1{Z_1} \). Similarly, for $Z_2$, the intrinsic graph is \( G_{int}^2 = {Z_2}^{t}\mathbb{L}_2{Z_2} \) and the penalty graph is \( G_{pen}^2 = {Z_2}^{t}\mathbb{U}_2{Z_2} \). Within the graph embedding (GE) framework, we employ the weighting scheme of Local Fisher Discriminant Analysis (LFDA) \cite{sugiyama2007dimensionality}. Consequently, the weights for the LFDA model's intrinsic and penalty graphs are determined as follows:

\begin{align}
\label{eq.25}
 \prescript{}{i}\Delta_{kl}^{int} & =   \begin{cases} 
    \frac{\lambda_{kl}}{N_{c_k}}, & c_k = c_l  \\
      0, & otherwise.
   \end{cases}\\
\label{eq.26}
 \prescript{}{i}\Delta_{kl}^{pen} & =   \begin{cases} 
    \lambda_{kl}(\frac{1}{N} - \frac{1}{N_{c_k}}), & c_k = c_l \\
      \frac{1}{N}, & otherwise.
   \end{cases}
\end{align}
Where $ i = 1 ,2.$
In this context, $N_{c_k}$ denotes the number of samples within the class labeled as $c_k$, while $ \lambda_{kl}$ quantifies the similarity between $z_k$ and $z_l$, where $z_k, z_l \in Z_i ~(i =1, 2) $. The kernel function is utilized in this paper to compute the similarity measure, i.e., $\lambda_{kl} = exp(\frac{-||z_k - z_l||^2}{2\sigma^2})$ where $\sigma$ is a scaling parameter.
\begin{algorithm}
    \caption{GRVFL-MV Model Algorithm}
    \label{Algorithm for GBLSTSVM.}
    \textbf{Input:} Training datasets $X_A$ and $X_B$.\\
    \textbf{Output:} GRVFL-MV model. \\
    \vspace{-0.5cm}
    \begin{algorithmic}[1]
        \State Given the values for $c_1, c_2, c_3, \theta_1, \theta_2,$ and $\rho$.
        \State Compute $Z_1$ and $Z_2$ using Equation (\ref{eq8}).
        \State Compute intrinsic and penalty graph weights using Equations (\ref{eq.25}) and (\ref{eq.26}) for $Z_1$ and $Z_2$.
        \State Calculate Laplacian matrices $\mathbb{L}_i = \mathbb{D}_i - \prescript{}{i}\Delta^{int}$ and $\mathbb{U}_i = \mathbb{L}_i^p = \mathbb{D}_i^p - \prescript{}{i}\Delta^{pen}$ for $Z_i ~(i= 1, 2).$
        \State Compute \( G_{int}^1 = {Z_1}^{t}\mathbb{L}_1{Z_1} \) and \( G_{pen}^1 = {Z_1}^{t}\mathbb{U}_1{Z_1} \) for $Z_1$, and \( G_{int}^2 = {Z_2}^{t}\mathbb{L}_2{Z_2} \) and \( G_{pen}^2 = {Z_2}^{t}\mathbb{U}_2{Z_2} \) for $Z_2$.
        \State Compute \( G_1 = (G_{pen}^{1})^{-1}G_{int}^1 \) and \( G_2 = (G_{pen}^2)^{-1}G_{int}^2 \).
        \State Use Equation (\ref{eq.24}) to calculate $\beta_1$ and $\beta_2$.
        \State Use test condition (\ref{eq:19}) to classify a new data point. 
    \end{algorithmic}
\end{algorithm} 
\subsection{Computational Complexity}
\label{sec-4}
We analyze the computational complexity of our proposed GRVFL-MV model in this section. For \( X_A \), computing the graph embedding (GE) matrix \( G_1 \) using the methodologies from \cite{iosifidis2015graph} which account for intrinsic as well as penalty graph structures results in a time complexity of \( \mathcal{O}((m + h_1)^3 + (m + h_1)^2l) \). Similarly, for \( X_B \), computing \( G_2 \) yields a complexity of \( \mathcal{O}((n + h_2)^3 + (n + h_2)^2l) \). Thus, the GE process incurs a total computational complexity of \( \mathcal{O}((n + h_1)^3 + (n + h_1)^2l) + \mathcal{O}((m + h_2)^3 + (m + h_2)^2l) \). To solve our model, we address (\ref{eq.24}). The complexity of this step is primarily governed by the inversion of a square matrix with an order of \( (m+n+h_1 +h_2) \), leading to a computational complexity of \( \mathcal{O}((m+n+h_1 +h_2)^3) \). Consequently, the total computational complexity of our proposed model becomes \( \mathcal{O}((n + h_1)^3 + (n + h_1)^2l) + \mathcal{O}((m + h_2)^3 + (m + h_2)^2l) + \mathcal{O}((m+n+h_1 +h_2)^3) \approx  \mathcal{O}((m+n+h_1 +h_2)^3)\).

\section{Experiments and Result Discussion}
\label{sec-5}
The performance of the proposed model has been evaluated against the baseline models: SVM2K \cite{farquhar2005two}, MvTSVM \cite{xie2015multi}, RVFLwoDL1 (RVFL without direct link \cite{huang2006extreme} on `$view-A$'),  RVFLwoDL2 (RVFL without direct link \cite{huang2006extreme} on `$view-B$'), RVFL1 (RVFL \cite{pao1994learning} on `$view-A$') and RVFL2 (RVFL \cite{pao1994learning} on `$view-B$'), and MVLDM \cite{hu2024multiview}. Theoretical comparisons of the proposed model against the baseline models are made in the supplementary Section S.I. The datasets used for the evaluation are UCI \cite{dua2017uci}, KEEL \cite{derrac2015keel}, Animal with Attributes (AwA)\footnote{\url{http://attributes.kyb.tuebingen.mpg.de}}, and Corel5K\footnote{\url{https://wang.ist.psu.edu/docs/related/}}. 

\subsection{Experimental Setup}
The experiments are carried out on a PC equipped with an Intel(R) Xeon(R) Gold 6226R processor clocked at 2.90GHz and 128 GB of RAM. The system runs on Windows 11 and uses Python 3.11. The dual of the QPP in SVM2K \cite{farquhar2005two} and MvTSVM \cite{xie2015multi} is solved using the ``QP solvers" function from the CVXOPT package. The dataset is randomly divided, allocating 70\% of the samples for training and 30\% for testing. Hyperparameters are optimized and validated using a five-fold cross-validation approach. The regulazation parameters $c_i$ $(i= 1,2,3),$ the graph regularization parameters  $\theta_j$ $(j=1,2),$ and the coupling parameter $\rho$ are tuned within the range $ \lbrace 10^{-5}, 10^{-4}, \cdots, 10^{5} \rbrace$. In our experiments, we have taken $c_1 = c_2 = c_3$ and $\theta_1 = \theta_2$.
All the hyperparameters of the baseline models were also taken within the same range. The hidden neurons $(h_l)$ varies as 3:20:203.

\subsection{Experiments on UCI and KEEL Datasets}
\begin{table}[htp]
\label{my-table-1}
\caption{Average accuracy and average rank of the baseline models and the proposed GRVFL-MV model over UCI and KEEL datasets.}
\resizebox{1\linewidth}{!}{
\begin{tabular}{ccccccccc} \hline
Model & SVM2K \cite{farquhar2005two} & MvTSVM \cite{xie2015multi} & RVFLwoDL1 \cite{huang2006extreme} & RVFLwoDL2 \cite{huang2006extreme} & RVFL1 \cite{pao1994learning} & RVFL2 \cite{pao1994learning} & MVLDM \cite{hu2024multiview}  & GRVFL-MV (proposed) \\ \hline
Average ACC & $76.70$ & $64.73$ & $82.99$ & $82.35$ & $83.59$ & $81.59$ & $79.88$ & $\mathbf{85.68}$ \\ \hline
Average Rank & $5.59$ & $7.71$ & $4.02$ & $4.21$ & $3.45$ & $4.38$ & $4.91$ & $\mathbf{1.74}$ \\ \hline
\end{tabular}}
\end{table}

Within this subsection, we delve into analyzing the statistical significance of the results obtained from our experiments, specifically concentrating on datasets sourced from UCI and KEEL repositories \cite{dua2017uci, derrac2015keel}. Through our assessment, we encompass a total of 29 datasets. Given the absence of inherent multi-view characteristics in the UCI and KEEL datasets, we designated the 95\% principal component extracted from the original data as `$view-B$', while the unaltered data itself acts as `$view-A$'.  In order to thoroughly compare the performance levels between our proposed graph random vector functional link based on the multi-view learning (GRVFL-MV) model and the baseline models with optimized hyperparameters, please refer to Table S.1 in the supplementary file. The outcomes illustrated in Table 1 in the paper underline the average accuracy (ACC) and average rank of each model across the UCI and KEEL datasets. The baseline models SVM2K, MvTSVM, RVFLwoDL1, RVFLwoDL2, RVFL1, RVFL2, and MVLDM achieves the average ACC of $76.70\%$, $64.73\%$, $82.99\%$, $82.35\%$, $83.59\%$, $81.59\%$, and $79.88\%$, respectively. The GRVFL-MV model that we put forward exhibited an outstanding average accuracy rate of $85.68\%$, surpassing the performance levels of the baseline models. The difference in average ACC of the proposed model and the baseline models lies approximately between $ 4\% - 20\%$. This showcases the superior generalization capabilities inherent in our proposed model compared to the baseline models.

In order to further assess the effectiveness of the proposed model, a ranking method is utilized. This method involves assigning a rank to each model for every dataset, where the model that performs the best is given the lowest rank, and the model that performs the worst is given the highest rank. The average rank for each model is then calculated by finding the mean of its ranks across all datasets. If there are a total of $N$ datasets, each evaluated with $\lambda$ models, the rank of the $p^{th}$ model on the $t^{th}$ dataset can be represented as ${s^{t}_p}$. Subsequently, the average rank of the $p^{th}$ model is calculated as ${R}_p = \frac{1}{N}\sum_{t=1}^{N}s^{t}_p$. The proposed GRVFL-MV model has been able to achieve an average rank of 1.74. On the other hand, the baseline models such as SVM2K, MvTSVM, RVFLwoDL1, RVFLwoDL2, RVFL1, RVFL2, and MVLDM have average ranks of 5.59, 7.71, 4.02, 4.21, 3.45, 4.38, and 4.91, respectively. These results clearly demonstrate the superiority of the proposed model over the baseline models.

In order to assess the statistical significance of the proposed model, the Friedman test is utilized as outlined in \cite{demvsar2006statistical}. This particular test is designed to pinpoint noteworthy variations among the models being compared by scrutinizing their average ranks. The underlying assumption of the null hypothesis is that all models showcase an identical average rank, indicating an equivalent level of performance. The Friedman test adheres to a chi-squared distribution denoted as $\chi^2_F$ with $(\lambda-1)$ degrees of freedom, which can be calculated using the formula: $\chi^2_F = \frac{12 N}{\lambda(\lambda +1)} \left[ \sum_{p} {{R}_p}^2 - \frac{\lambda(\lambda+1)^2}{4} \right]$. Here, ${R}_p$ signifies the average rank of the $p$th model, $\lambda$ represents the total number of models, and $N$ denotes the number of datasets involved in the analysis. The Friedman statistic $F_F$ is determined by the formula: $ F_F = \frac{(N-1)\chi^2_F}{N(\lambda-1)-\chi^2_F}$, where the $F$-distribution is characterized by $(\lambda-1)$ and $(\lambda-1)\times(N-1)$ degrees of freedom. Given that $N = 29$ and $\lambda = 8$, the values of $\chi^2_F$ and $F_F$ are calculated to be 100.5 and 27.45, respectively. By consulting the $F$-distribution table at a 5\% level of significance $(\alpha)$, the critical value of $F_F(7, 196) = 2.05$ is obtained. Since the calculated value of $F_F$ is 27.45, which exceeds the critical value of 2.05, the null hypothesis is rejected. This outcome underscores a substantial statistical distinction among the models under comparison.

\begin{table}[htp]
\caption{The statistical comparison of the proposed GRVFL-MV model with the baseline models on UCI and KEEL datasets using the Nemenyi post hoc test.}
\label{table-2}
\resizebox{1\linewidth}{!}{
\begin{tabular}{lccccccc} \hline
& SVM2K \cite{farquhar2005two} & MvTSVM \cite{xie2015multi} & RVFLwoDL1 \cite{huang2006extreme} & RVFLwoDL2 \cite{huang2006extreme} & RVFL1 \cite{pao1994learning} & RVFL2 \cite{pao1994learning} & MVLDM \cite{hu2024multiview}  \\ \hline
\multicolumn{1}{c}{GRVFL-MV (proposed)} & \cmark & \cmark & \cmark & \cmark & \xmark & \cmark & \cmark \\ \hline
\multicolumn{7}{l}{\begin{tabular}[c]{@{}l@{}} \cmark~ indicates that the model listed in the row is superior to the model mentioned in the column, while \xmark~indicates otherwise.\end{tabular}}
\end{tabular}}
\end{table}

In addition, the Nemenyi post hoc test \cite{demvsar2006statistical} is utilized to delve deeper into the differences between the models. If the average ranks of the models exhibit a variance of at least the critical difference ($CD$), then they are deemed to be significantly distinct. The $CD$ is determined by the formula: $ CD = q_\alpha {\begin{bmatrix} \frac{\lambda(\lambda+1)}{6N} \end{bmatrix}}^{1/2}$. On calculating, we get the $CD = 1.94$. The results presented in Table \ref{table-2} provide clear evidence that the average rank difference between the baseline model and the proposed GRFVL-MVL is larger than the critical difference ($CD$) value. To be more precise, the average rank difference between MvTSVM and GRVFL-MV is $5.97$, which is greater than the $CD$ value of $1.94$. Similarly, the average rank difference between MVLDM and GRVFL-MV is $3.17$, also surpassing the $CD$ value. Therefore, these results unequivocally demonstrate that the proposed model exhibits significant dissimilarity compared to the baseline models.

Furthermore, a pairwise win-tie-loss sign test is conducted under the assumption of equal performance between the two models under the null hypothesis. It is anticipated that each model will emerge victorious in roughly half of the total datasets, denoted as $\frac{N}{2}$, where $N$ represents the total number of datasets. To establish statistical significance, a model must secure wins in approximately $\frac{N}{2} + 1.96\frac{\sqrt{N}}{2}$ datasets more than its counterpart. In scenarios where there is an even number of ties between the models, the ties are evenly distributed. Conversely, in cases of an odd number of ties, one tie is discounted, and the remaining ties are divided between the models. With $N$ set at 29, a minimum of 19.78 wins is required to establish a significant difference between the two models. 
\begin{table*}[htp]
\caption{Pairwise win-tie-loss test of proposed GRVFL-MV model and baseline models on UCI and KEEL datasets.}
\label{table-3}
\resizebox{1\linewidth}{!}{
\begin{tabular}{lccccccc} \hline
\multicolumn{1}{l}{} & SVM2K  \cite{farquhar2005two} & MvTSVM \cite{xie2015multi} & RVFLwoDL1 \cite{huang2006extreme} & RVFLwoDL2 \cite{huang2006extreme} & \multicolumn{1}{c}{RVFL1} \cite{pao1994learning} & \multicolumn{1}{c}{RVFL2} \cite{pao1994learning} & \multicolumn{1}{c}{Large\_Margin} \cite{hu2024multiview}  \\ \hline
MvTSVM \cite{xie2015multi} & {[}2, 2, 25{]} & \multicolumn{1}{l}{} & \multicolumn{1}{l}{} & \multicolumn{1}{l}{} &  &  &  \\
RVFLwoDL1 \cite{huang2006extreme} & {[}20, 1, 8{]} & {[}29, 0, 0{]} & \multicolumn{1}{l}{} & \multicolumn{1}{l}{} &  &  &  \\
RVFLwoDL2 \cite{huang2006extreme} & {[}20, 2, 7{]} & {[}28, 0, 1{]} & {[}11, 4, 14{]} & \multicolumn{1}{l}{} &  &  &  \\
RVFL1 \cite{pao1994learning} & {[}23, 1, 5{]} & {[}29, 0, 0{]} & {[}13, 10, 6{]} & {[}15, 4, 10{]} &  &  &  \\
RVFL2 \cite{pao1994learning} & {[}20, 1, 8{]} & {[}27, 1, 1{]} & {[}10, 5, 14{]} & {[}11, 9, 9{]} & \multicolumn{1}{c}{{[}7, 7, 15{]}} &  &  \\
Large\_Margin \cite{hu2024multiview} & {[}16, 1, 12{]} & {[}26, 0, 3{]} & {[}11, 0, 18{]} &{[}11, 0, 18{]} & \multicolumn{1}{c}{{[}9, 0, 20{]}} & \multicolumn{1}{c}{{[}13, 0, 16{]}} &  \\
GRVFL-MV (proposed) & {[}27, 0, 2{]} & {[}29, 0, 0{]} & {[}24, 1, 4{]} & {[}25, 1, 3{]} & {[}22, 2, 5{]} & \multicolumn{1}{c}{{[}26, 1, 2{]}} & \multicolumn{1}{c}{{[}27, 0, 2{]}} \\ \hline
\end{tabular}} \begin{flushleft}
wherein
    $\begin{bmatrix}
    x & y & z
    \end{bmatrix}$, $x$ signifies no. of wins, $y$ no. of draws, and $z$ no. of losses.
\end{flushleft}
\end{table*}


{The results depicted in Table \ref{table-3} showcase a distinct advantage for our proposed model over the baseline model. Clearly, we can see that our proposed model wins on at least 22 ($>$ 19.78) datasets out of 27 datasets. Hence, our proposed model is far superior to the baseline models. In particular, we can observe that our proposed GRVFL-MV surpasses SVM2K in 27 datasets out of the 29 datasets. Furthermore, it outperforms the MvTSVM in all 29 datasets while excelling against MVLDM  in 27 datasets out of the 29. The findings unequivocally affirm that our proposed model exhibits a significantly higher performance level compared to the baseline models.}

\subsection{Experiments on Corel5k Datasets}
The  Corel5k\footnote{\url{https://wang.ist.psu.edu/docs/related/}}
dataset serves as a prominent benchmark in the realms of computer vision and image processing. Comprising 5,000 images across 50 distinct categories, each category includes 100 images. This dataset is widely used for a variety of applications, including content-based image retrieval (CBIR), object recognition, and image classification. We divide the dataset into 50 binary datasets using the one-versus-rest approach. For every binary dataset, 100 photographs from the target category make up the positive class, while another 100 images are chosen at random from the other categories to make up the negative class. This approach facilitates the creation of tailored binary datasets for focused analysis and experimentation.
\begin{table}[ht!]
\caption{Average accuracy and average rank of the baseline models and the proposed model GRVFL-MV over Corel5k datasets.}
\label{table-4}
\resizebox{1\linewidth}{!}{
\begin{tabular}{ccccccccc} \hline
Model & SVM2K \cite{farquhar2005two} & MvTSVM \cite{xie2015multi} & RVFLwoDL1 \cite{huang2006extreme} & RVFLwoDL2 \cite{huang2006extreme} & RVFL1 \cite{pao1994learning} & RVFL2 \cite{pao1994learning} & MVLDM \cite{hu2024multiview} & GRVFL-MV (proposed) \\ \hline
Average ACC & 74.87 & 49.93 & 74.98 & 74.83 & 76.33 & 75.43 & 69.87 & 77.33 \\ \hline
Average Rank & 4.11 & 7.91 & 3.99 & 3.97 & 3.42 & 4.05 & 5.61 & 2.94 \\ \hline
\end{tabular}}
\end{table}

The results presented in Table \ref{table-4} provide clear evidence that our proposed model outperforms the baseline models. With an average accuracy (ACC) of 77.33\%, our model achieves the highest accuracy among all the models compared. Additionally, the proposed GRVFL-MV model obtains the lowest average rank of 2.94, indicating its superior performance compared to the baseline models. These findings are from the 50 datasets from Corel5k, which further solidifies the superiority of our proposed model. For more detailed experimental results, please refer to Table S.2 in the supplementary file.

\subsection{Experiments on Animal with Attributes (AwA) Datasets}
AwA\footnote{\url{http://attributes.kyb.tuebingen.mpg.de}} is a large dataset with 30,475 images covering 50 different animal types. Six representations of pre-extracted features are used to describe each image. For our analysis, we focus on a subset of ten specific test classes drawn from this dataset. These classes include animals like chimpanzees, giant pandas, leopards, Persian cats, pigs, hippopotamuses, humpback whales, raccoons, rats, and seals, totaling 6180 images. For our analysis, we employ two distinct feature representations: a 2000-dimensional $L_1$ normalized speeded-up robust features (SURF) descriptors ($view-B$) and a 252-dimensional histogram of oriented gradient features ($view-A$). We build and train 45 binary classifiers using the one-against-one method for every possible combination of class pairs in the dataset.
\begin{table}[ht!]
\caption{Average accuracy and average rank of the baseline models and the proposed model GRVFL-MV over AwA datasets.}
\label{table-5}
\resizebox{1\linewidth}{!}{
\begin{tabular}{ccccccccc} \hline
Model & SVM2K \cite{farquhar2005two} & MvTSVM \cite{xie2015multi} & RVFLwoDL1 \cite{huang2006extreme} & RVFLwoDL2 \cite{huang2006extreme} & RVFL1 \cite{pao1994learning} & RVFL2 \cite{pao1994learning} & MVLDM \cite{hu2024multiview} & GRVFL-MV (proposed) \\ \hline
Avearge ACC & 77.46 & 64.31 & 71.74 & 75.99 & 72.87 & 77.46 & 73.33 & 83.29 \\ \hline
Avearge Rank & 3.46 & 6.84 & 5.99 & 4.32 & 5.22 & 3.59 & 5.02 & 1.56 \\ \hline 
\end{tabular}}
\end{table}
The average rank and ACC of our proposed model are displayed in Table \ref{table-5} in relation to the baseline models. Across the 45 datasets from AwA, our model achieved the highest average accuracy and the lowest average rank. The average ACC of proposed  GRVFL-MV is 83.29, which is upto 4\% to 20\% higher than the baseline models. Also, GRVFL-MV has the lowest average rank of 1.56. The results clearly demonstrate the superior performance of the proposed GRVFL-MV model over the baseline models. For more detailed experimental findings, please refer to Table S.3 in the supplementary file.

\subsection{Sensitivity Analysis of Hyperparameters \texorpdfstring{$C_1$}{C1} and \texorpdfstring{$C_2$}{C2}}
\begin{figure*}[ht!]
\begin{minipage}{.5\linewidth}
\centering
\subfloat[bupa or liver-disorders]{\label{fig:1a}\includegraphics[scale=0.35]{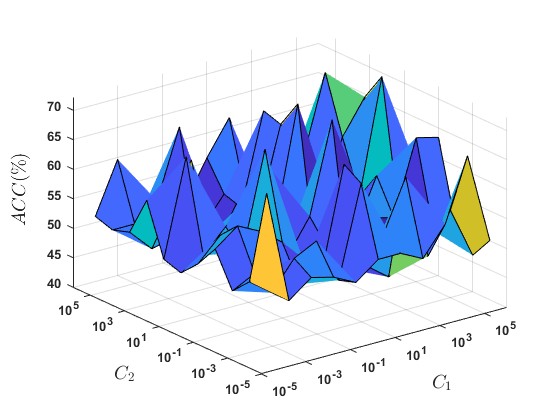}}
\end{minipage}
\begin{minipage}{.5\linewidth}
\centering
\subfloat[hepatitis]{\label{fig:1b}\includegraphics[scale=0.35]{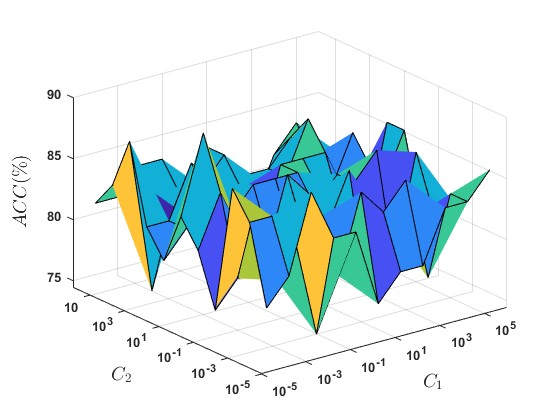}}
\end{minipage}
\begin{minipage}{.5\linewidth}
\centering
\subfloat[ilpd\_indian\_liver]{\label{fig:1d}\includegraphics[scale=0.35]{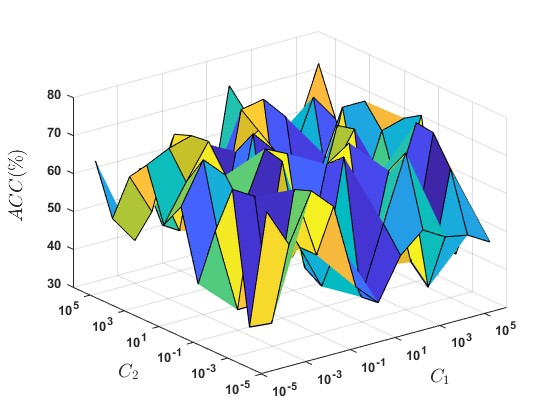}}
\end{minipage}
\begin{minipage}{.5\linewidth}
\centering
\subfloat[monks\_2]{\label{fig:1e}\includegraphics[scale=0.35]{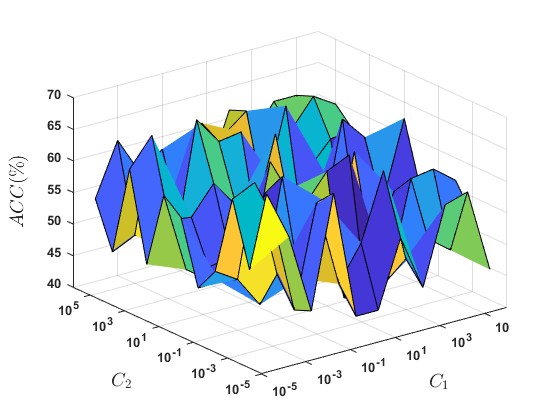}}
\end{minipage}
\caption{The effect of hyperparameter $(c_1, c_2)$ tuning on the accuracy (ACC) of some UCI and KEEL datasets on the performance of proposed GRVFL-MV model.}
\label{}
\end{figure*}
To gain a comprehensive understanding of how hyperparameters affect the generalization ability of the proposed GRVFL-MV model,  we conducted a systematic exploration of the hyperparameter space by tuning the values of $c_1$ and $c_2$. This helps us to identify the optimal configuration that maximizes predictive accuracy and improves the model's resilience to previously unseen data. Fig. 2.  visually represents how the accuracy of the model behaves when the hyperparameters are tuned. The visual clearly shows that the proposed model is highly sensitive to the values of hyperparameters $c_1$ and $c_2$. In Fig. 2. (a), we see that optimal accuracy is achieved when $c_1 = 10^{5}$ and $c_2 = 10 ^{4}$ whereas, in Fig. 2. (c), optimal configuration is found at two distinct coordinates, $(10^{5}, 10^{5})$ and $(10, 10^{5})$.  Similar observations can be made for Fig. 2. (b) and Fig. 2. (d). Overall, these findings underscore the need for careful selection of hyperparameter values to achieve optimal model performance.

\subsection{Sensitivity Analysis of Coupling Parameter \texorpdfstring{$\rho$}{rho}}

The primal optimization problem (\ref{eq:1}) combines two distinct classification objectives, each corresponding to a different view. These objectives are linked through a coupling term ${\xi_1^{t}}\xi_2$, where $\rho$ serves as the regularization constant known as the coupling parameter. We analyzed the impact of $\rho$ on our model's performance by fixing other parameters at their optimal values and tuning $\rho$ within the specified range outlined in the experimental setup. 
\begin{figure*}[ht!]
\begin{minipage}{.32\linewidth}
\centering
\subfloat[UCI and KEEL]{\label{fig:2a}\includegraphics[scale=0.25]{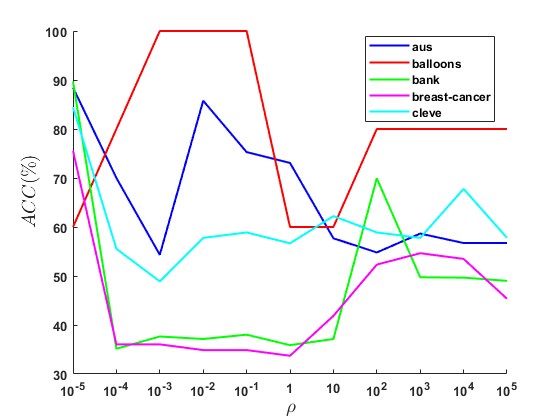}}
\end{minipage}
\begin{minipage}{.32\linewidth}
\centering
\subfloat[AwA]{\label{fig:2b}\includegraphics[scale=0.25]{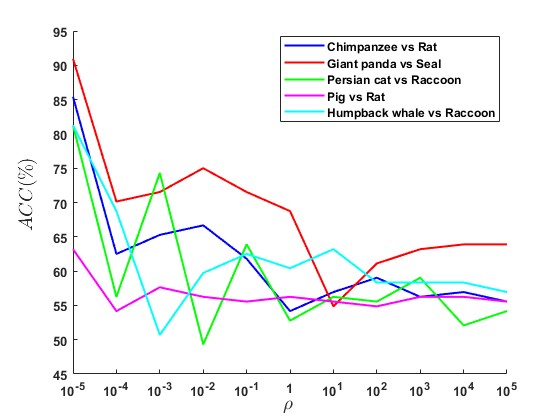}}
\end{minipage}
\begin{minipage}{.25\linewidth}
\centering
\subfloat[Corel5k]{\label{fig:2d}\includegraphics[scale=0.25]{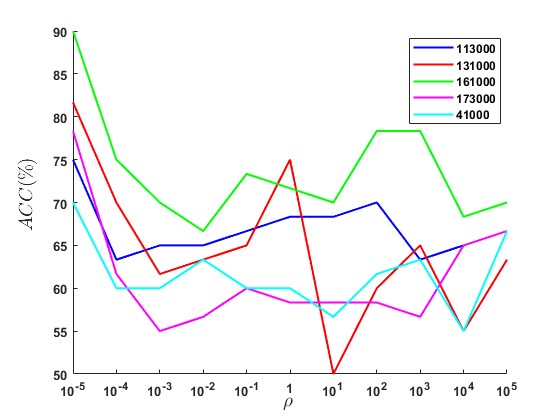}}
\end{minipage}
\caption{The effect of coupling parameter $\rho$ tuning on the accuracy (ACC) of UCI and KEEl, AwA, and Corel5k datasets on the performance of proposed GRVFL-MV model.}
\label{fig:hyperparameter-analysis}
\end{figure*}

The performance of the proposed model across different datasets is depicted in Fig. 3. It can be observed from Fig. 3. (a) that the proposed GRVFL-MV model achieves the highest accuracy when $\rho$ is set to $10^{-5}$. This trend is also evident in Fig.3. (b) and Fig.3. (c). These results indicate that the optimal effect of coupling terms and improved generalization performance are achieved when $\rho$ is tuned to $10^{-5}$.  

\subsection{Sensitivity Analysis of Graph Regularization Parameter \texorpdfstring{$\theta$}{theta}}
The primal optimization problem (\ref{eq:1}) involves two graph regularization parameters $\theta_1$ and $\theta_2$ for each view, with the goal of preserving the geometrical aspects of multiview data through the graph embedding GE) framework in the model. In our experiment, we set $\theta_1 = \theta_2 = \theta$ to study the effect of geometrical properties of the multiview data through the LFDA technique under the GE framework.
\begin{figure*}[ht!]
\label{theta}
\begin{minipage}{.5\linewidth}
\centering
\subfloat[Corel5k]{\label{fig:3b}\includegraphics[scale=0.35]{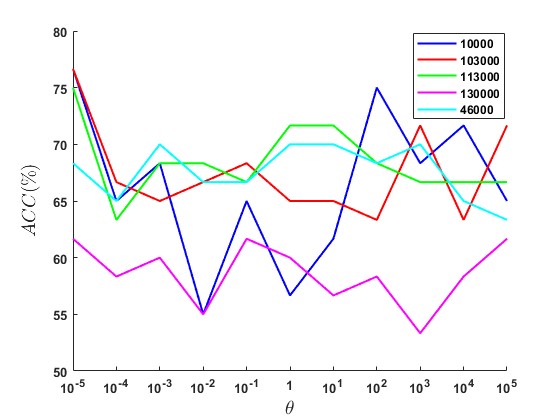}}
\end{minipage}
\begin{minipage}{.5\linewidth}
\centering
\subfloat[AwA]{\label{fig:3d}\includegraphics[scale=0.35]{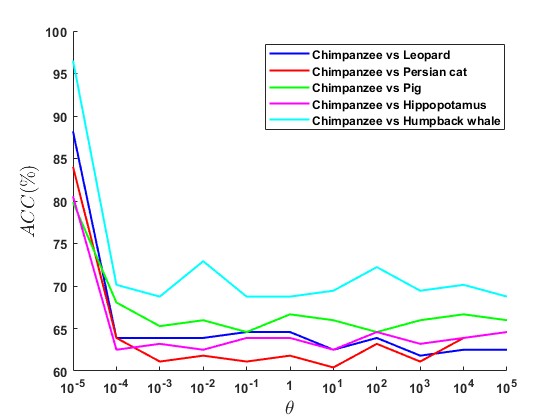}}
\end{minipage}
\caption{The effect of graph embedding  parameter $\theta$ tuning on the accuracy (ACC) of Corel5k and AwA datasets on the performance of proposed GRVFL-MV model.}
\label{fig:hyperparameter-analysis1}
\end{figure*}
The impact of tuning $\theta$ on the efficacy of the GRVFL-MV model is illustrated in Fig. 4. In Fig. 4. (a) when utilizing datasets from Corel5k, it is evident that the model's performance is most favorable initially at the minimum $\rho$ value, specifically at $\rho = 10^{-5}$, and subsequently performance increases for $\rho > 1$. Conversely, in Fig.  4. (b), the optimal performance is achieved at $\rho = 10^{-5}$ but then experiences a significant decline. These results emphasize the critical importance of selecting the appropriate $\rho$ value for maximizing the proposed model's performance.

\section{{Comprehensive Performance Analysis Across Dataset Types}}
\label{comprehensive analysis}
{In this section, we offer an extensive evaluation of the proposed model's performance across a diverse range of datasets used in our experiments. We classify these datasets into five categories: high-dimensional, medium-dimensional, low-dimensional, sparse, and imbalanced datasets. This analysis aims to provide a thorough understanding of how the model performs under different conditions and dataset characteristics.}

\subsection{ {\textbf{Performance on High-Dimensional Datasets}} } 

\begin{table}[ht!]
\caption{{High-dimensional datasets (More than 30 Features) with imbalance ratios}}
\centering
\resizebox{\textwidth}{!}{%
\begin{tabular}{|l|c|c|c|c|c|}
\hline
\textbf{{Dataset Name}} & \textbf{{No. of Samples}} & \textbf{{No. of Features}} & \textbf{{Positive Class}} & \textbf{{Negative Class}} & \textbf{{Imbalance Ratio (IR)}} \\ \hline
{breast\_cancer\_wisc\_diag} & {569} & {30} & {212} & {357} & {0.594} \\ \hline
{breast\_cancer\_wisc\_prog} & {198} & {33} & {47} & {151} & {0.311} \\ \hline
{conn\_bench\_sonar\_mines\_rocks} & {208} & {60} & {97} & {111} & {0.874} \\ \hline
{hill\_valley} & {1,212} & {100} & {606} & {606} & {1.000} \\ \hline
{oocytes\_merluccius\_nucleus\_4d} & {1,022} & {41} & {337} & {685} & {0.492} \\ \hline
{cylinder\_bands} & {512} & {35} & {200} & {312} & {0.641} \\ \hline
{chess\_krvkp} & {3,196} & {36} & {1,527} & {1,669} & {0.915} \\ \hline
{musk\_1}       & {476}   & {166} & {207} & {269} & {0.7695} \\ \hline
{musk\_2}       & {6,598} & {166} & {1,017} & {5,581} & {0.1822} \\ \hline
\end{tabular}%
}
\label{table:high_dimensional_datasets_with_imbalance_ratios}
\end{table}

\begin{table}[ht!]
\caption{{Accuracy comparison of GRVFL-MV and baseline models on high-dimensional datasets}}
\centering
\resizebox{\textwidth}{!}{%
\begin{tabular}{|l|c|c|c|c|c|c|c|c|c|}
\hline
\textbf{{Dataset Name}} & \textbf{{SVM2K (\%)}} & \textbf{{MvTSVM (\%)}} & \textbf{{RVFLwoDL1 (\%)}} & \textbf{{RVFLwoDL2 (\%)}} & \textbf{{RVFL1 (\%)}} & \textbf{{RVFL2 (\%)}} & \textbf{{MVLDM (\%)}} & \textbf{{GRVFL-MV (\%)}} \\ \hline
{breast\_cancer\_wisc\_diag} & {95.49} & {88.60} & {92.42} & {97.08} & {95.83} & {96.49} & {93.15} & {98.57} \\ \hline
{breast\_cancer\_wisc\_prog} & {58.33} & {58.33} & {68.33} & {68.33} & {73.33} & {68.33} & {71.17} & {75.00} \\ \hline
{conn\_bench\_sonar\_mines\_rocks} & {80.95} & {46.03} & {80.54} & {74.60} & {88.54} & {73.02} & {75.81} & {77.78} \\ \hline
{hill\_valley} & {60.98} & {53.30} & {68.05} & {51.92} & {68.78} & {51.92} & {56.20} & {77.75} \\ \hline
{oocytes\_merluccius\_nucleus\_4d} & {74.27} & {64.82} & {82.41} & {81.11} & {83.71} & {80.78} & {75.16} & {83.39} \\ \hline
{cylinder\_bands} & {68.18} & {60.39} & {76.62} & {74.03} & {72.08} & {74.68} & {71.90} & {74.68} \\ \hline
{chess\_krvkp} & {80.45} & {82.35} & {95.62} & {93.33} & {95.62} & {94.68} & {97.70} & {96.98} \\ \hline
{musk\_1}       & {76.19} & {46.85} & {79.72} & {87.41} & {83.22} & {76.22} & {75.81} & {87.41} \\ \hline
{musk\_2}       & {78.46} & {14.75} & {95.15} & {95.00} & {96.16} & {95.71} & {83.43} & {96.16} \\ \hline
\end{tabular}%
}
\label{table:high_dimensional_accuracy_comparison}
\end{table}

    {Datasets with more than \(30\) features have been classified as high-dimensional, as detailed in Table \ref{table:high_dimensional_datasets_with_imbalance_ratios}. The proposed GRVFL-MV model demonstrates outstanding performance on high-dimensional datasets, achieving impressive accuracies of 77.75\% on \textit{hill\_valley} (100 features) and 98.57\% on \textit{breast\_cancer\_wisc\_diag} (30 features), outperforming all baseline models. High-dimensional datasets typically pose significant challenges due to the large number of features compared to the number of samples, leading to issues like overfitting or inefficient feature utilization in models like SVM2K and MvTSVM. In contrast, GRVFL-MV leverages graph-based learning to uncover meaningful structural patterns within multiview data. By incorporating graph regularization, it ensures robust feature selection, improves generalization, and effectively mitigates the impact of dimensionality-induced noise and redundancy. Table \ref{table:high_dimensional_accuracy_comparison} provides a detailed comparison of classification accuracies across baseline models and the proposed GRVFL-MV, demonstrating its superior performance on high-dimensional datasets.}

\subsection{ {\textbf{Effectively Handling Imbalanced Datasets}} } 

\begin{figure}[ht!]
       \centering
        \includegraphics[width=9cm, height=6cm]{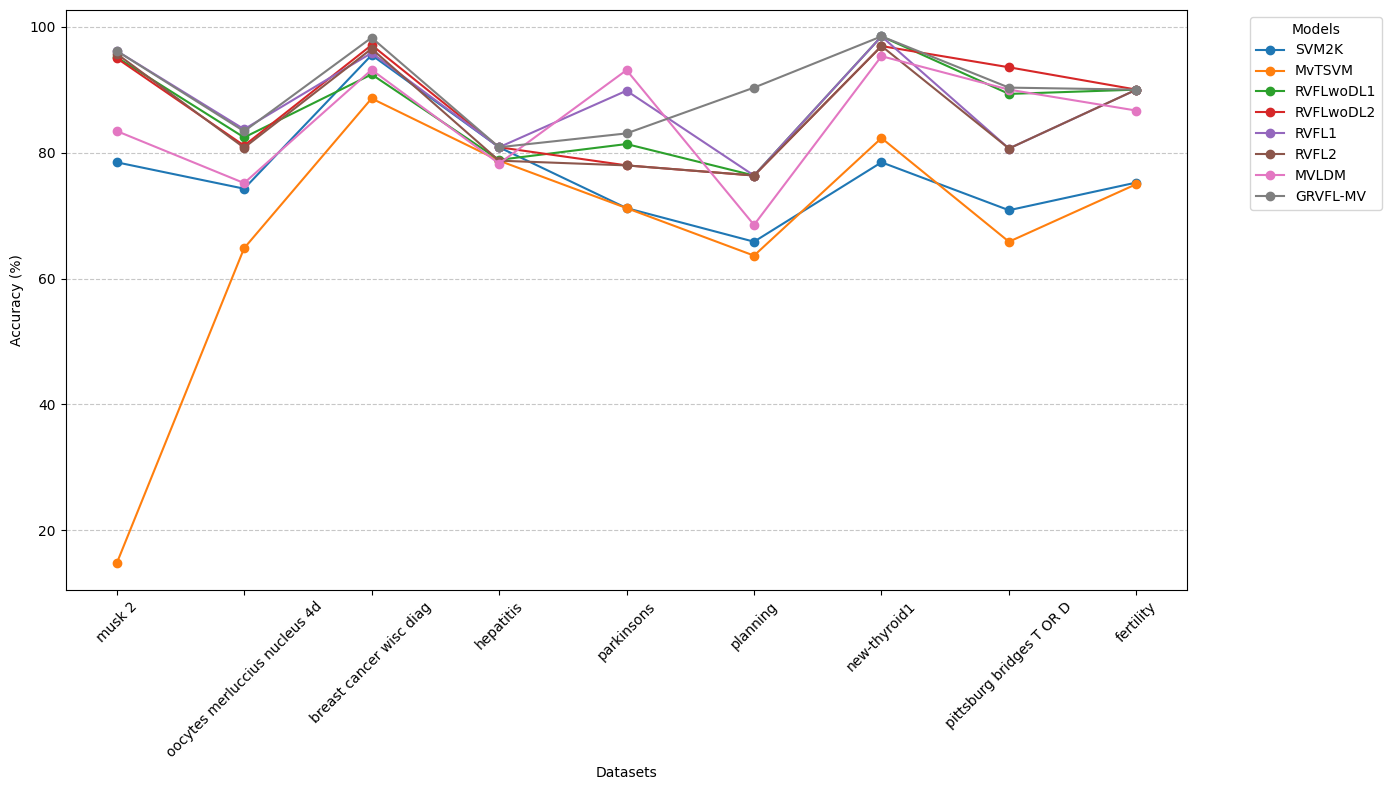}
        \caption{Accuracy of Different Models on Highly Imbalanced Datasets.}
        \label{fig:imbalance}
\end{figure}

{Datasets with significant class imbalance, such as \textit{breast\_cancer\_wisc\_diag} (IR = 0.594) and \textit{musk\_2} (IR = 0.1822), often pose challenges to traditional classification models. However, GRVFL-MV consistently demonstrates exceptional performance, achieving accuracies of 98.57\% and 96.16\%, respectively. By leveraging complementary multiview information, GRVFL-MV effectively balances the contributions of both majority and minority classes during training. Unlike simpler models such as RVFLwoDL1 and RVFL1, the proposed model mitigates bias introduced by class imbalance through its robust multiview alignment, ensuring that all classes are equitably represented within the learned latent space. Figure \ref{fig:imbalance} provides a comparative visualization of the performance of GRVFL-MV and baseline models across highly imbalanced datasets, showcasing its superior ability to address these challenges.}

\subsection{ {\textbf{Sparse and High-Dimensional Datasets}} } 

\begin{table}[ht!]
\caption{{Sparse datasets with imbalance ratios (IR)}}
\centering
\resizebox{\textwidth}{!}{
\begin{tabular}{|l|c|c|c|c|c|}
\hline
\textbf{{Dataset Name}} & \textbf{{No. of Samples}} & \textbf{{No. of Features}} & \textbf{{Positive Class}} & \textbf{{Negative Class}} & \textbf{{Imbalance Ratio (IR)}} \\ \hline
{hill\_valley}  & {1,212} & {100} & {606} & {606} & {1.000} \\ \hline
{musk\_1}       & {476}   & {166} & {207} & {269} & {0.7695} \\ \hline
{musk\_2}       & {6,598} & {166} & {1,017} & {5,581} & {0.1822} \\ \hline
\end{tabular}
}
\label{table:sparse_datasets_with_ratios}
\end{table}

\begin{figure}[ht!]
       \centering
        \includegraphics[width=9cm, height=6cm]{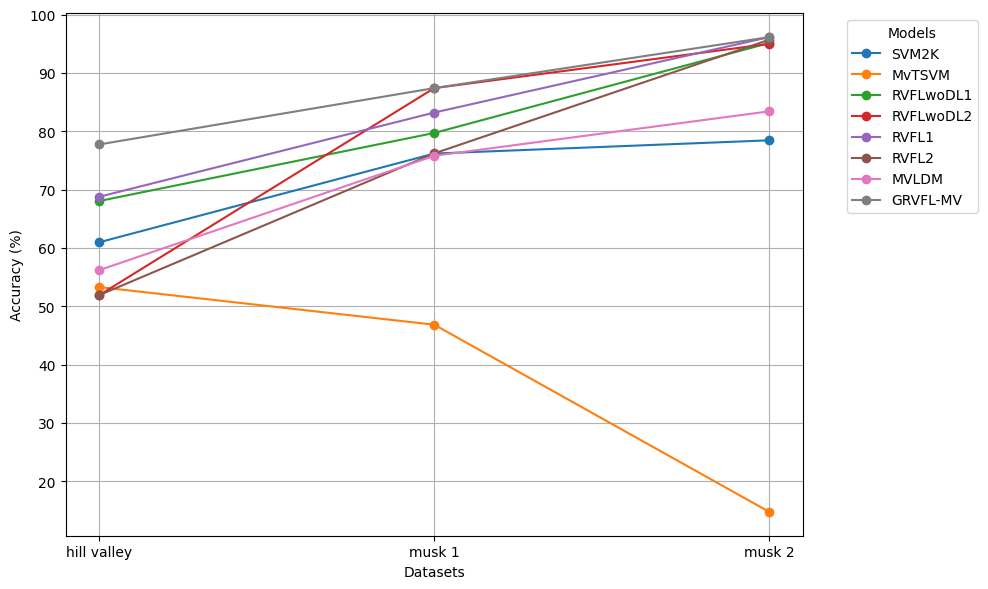}
        \caption{Accuracy of Different Models on Spare Datasets.}
        \label{fig:spare}
\end{figure}

\begin{table}[htp]
\caption{{Accuracy comparison of GRVFL-MV and baseline models on sparse datasets}}
\centering
\resizebox{\textwidth}{!}{
\begin{tabular}{|l|c|c|c|c|c|c|c|c|}
\hline
\textbf{{Dataset Name}} & \textbf{{SVM2K (\%)}} & \textbf{{MvTSVM (\%)}} & \textbf{{RVFLwoDL1 (\%)}} & \textbf{{RVFLwoDL2 (\%)}} & \textbf{{RVFL1 (\%)}} & \textbf{{RVFL2 (\%)}} & \textbf{{MVLDM (\%)}} & \textbf{{GRVFL-MV (\%)}} \\ \hline
{hill\_valley} & {60.98} & {53.30} & {68.05} & {51.92} & {68.78} & {51.92} & {56.20} & {77.75} \\ \hline
{musk\_1} & {76.19} & {46.85} & {79.72} & {87.41} & {83.22} & {76.22} & {75.81} & {87.41} \\ \hline
{musk\_2} & {78.46} & {14.75} & {95.15} & {95.00} & {96.16} & {95.71} & {83.43} & {96.16} \\ \hline
\end{tabular}
}
\label{table:sparse_datasets_accuracy_comparison}
\end{table}

    {Following the methodology of \cite{shi2021random} for the selection of sparse datasets, we consider high-dimensional sparse datasets, namely, \textit{musk\_1} (166 features), \textit{musk\_2} (166 features), and \textit{hill\_valley} (100 features) to evaluate the performance of the proposed GRVFL-MV against the baseline model as listed out in Table \ref{table:sparse_datasets_with_ratios}. These datasets pose significant challenges due to their high dimensionality and sparsity of features. The proposed GRVFL-MV model demonstrates remarkable performance on these datasets, showcasing its ability to address these complexities effectively. For \textit{musk\_1}, GRVFL-MV achieves an accuracy of 87.41\%, outperforming most baseline models and highlighting its robustness in handling datasets with an imbalance ratio (IR) of 0.7695. On \textit{musk\_2}, which features an even lower IR of 0.1822, GRVFL-MV achieves a notable accuracy of 96.16\%, demonstrating its capability to manage highly sparse and imbalanced datasets by leveraging graph-based learning to capture meaningful inter-feature relationships. Similarly, on the \textit{hill\_valley} dataset, despite its noisy and sparse feature set, GRVFL-MV achieves 77.75\% accuracy, outperforming traditional models like SVM2K and MvTSVM. This consistent performance across all three datasets underscores GRVFL-MV's effectiveness in handling sparsity and high dimensionality by integrating multiview information and employing graph regularization to emphasize key data patterns. Figure \ref{fig:spare} and Table \ref{table:sparse_datasets_accuracy_comparison} offer a comprehensive comparison of the proposed GRVFL-MV model against all baseline models on sparse datasets, effectively showcasing its superior performance and competitive accuracy.}

\subsection{ {\textbf{Performance on Medium-dimensional Datasets}}}  


\begin{table}[ht!]
\caption{{Medium-dimensional datasets (10–30 Features) with imbalance ratios}}
\centering
\resizebox{\textwidth}{!}{%
\begin{tabular}{|l|c|c|c|c|c|}
\hline
\textbf{{Dataset Name}} & \textbf{{No. of Samples}} & \textbf{{No. of Features}} & \textbf{{Positive Class}} & \textbf{{Negative Class}} & \textbf{{Imbalance Ratio (IR)}} \\ \hline
{bank} & {4,521} & {16} & {521} & {4,000} & {0.130} \\ \hline
{aus} & {690} & {15} & {307} & {383} & {0.802} \\ \hline
{cleve} & {297} & {13} & {137} & {160} & {0.856} \\ \hline
{parkinsons} & {195} & {22} & {48} & {147} & {0.327} \\ \hline
{planning} & {182} & {12} & {52} & {130} & {0.400} \\ \hline
{oocytes\_trisopterus\_nucleus\_2f} & {912} & {25} & {385} & {527} & {0.731} \\ \hline
{checkerboard\_Data} & {690} & {14} & {307} & {383} & {0.802} \\ \hline
{hepatitis} & {155} & {19} & {33} & {124} & {0.266} \\ \hline
\end{tabular}%
}
\label{table:medium_dimensional_datasets_with_ratios}
\end{table}

\begin{table}[ht!]
\caption{{Accuracy comparison of GRVFL-MV and baseline models on medium-dimensional datasets}}
\centering
\resizebox{\textwidth}{!}{%
\begin{tabular}{|l|c|c|c|c|c|c|c|c|}
\hline
\textbf{{Dataset Name}} & \textbf{{SVM2K (\%)}} & \textbf{{MvTSVM (\%)}} & \textbf{{RVFLwoDL1 (\%)}} & \textbf{{RVFLwoDL2 (\%)}} & \textbf{{RVFL1 (\%)}} & \textbf{{RVFL2 (\%)}} & \textbf{{MVLDM (\%)}} & \textbf{{GRVFL-MV (\%)}} \\ \hline
{bank} & {80.74} & {71.86} & {85.54} & {85.39} & {85.83} & {85.61} & {73.67} & {89.98} \\ \hline
{aus} & {87.02} & {71.15} & {85.98} & {86.06} & {85.98} & {85.98} & {71.98} & {87.50} \\ \hline
{cleve} & {80.00} & {75.56} & {80.00} & {85.56} & {81.11} & {81.11} & {84.27} & {84.44} \\ \hline
{parkinsons} & {71.19} & {71.19} & {81.36} & {77.97} & {89.83} & {77.97} & {93.10} & {83.05} \\ \hline
{planning} & {65.85} & {63.64} & {76.36} & {76.36} & {76.36} & {76.36} & {68.52} & {90.32} \\ \hline
{oocytes\_trisopterus\_nucleus\_2f} & {78.83} & {58.39} & {86.13} & {78.83} & {85.04} & {78.83} & {82.05} & {84.67} \\ \hline
{checkerboard\_Data} & {87.02} & {43.75} & {86.98} & {86.06} & {86.98} & {86.98} & {84.06} & {87.50} \\ \hline
{hepatitis} & {80.85} & {78.72} & {78.85} & {80.85} & {80.85} & {78.72} & {78.26} & {80.85} \\ \hline
\end{tabular}%
}
\label{table:medium_dimensional_accuracy_comparison}
\end{table}

    { Datasets with more than $10$ and fewer than $30$ features are categorized as medium-dimensional datasets. Table \ref{table:medium_dimensional_datasets_with_ratios} provides a comprehensive list of all UCI and KEEL datasets classified as medium-dimensional for this experiment. Table \ref{table:medium_dimensional_accuracy_comparison} presents a detailed performance comparison between the baseline models and the proposed GRVFL-MV model on medium-dimensional datasets. These datasets often strike a balance between feature dimensionality and sample size, introducing unique challenges for classification models. GRVFL-MV exhibits robust performance on datasets such as \textit{bank} (16 features) and \textit{cleve} (13 features), achieving accuracies of 89.98\% and 84.44\%, respectively. The model effectively leverages complementary multiview features and utilizes graph-based regularization to ensure consistent classification performance. On more challenging datasets, such as \textit{parkinsons} (22 features), which are impacted by class imbalance, GRVFL-MV achieves an accuracy of 83.05\%, significantly outperforming simpler models like RVFL1 and SVM2K.} 

\subsection{ {\textbf{Performance on Low-dimensional Datasets}}}


\begin{table}[ht!]
\caption{{Low-dimensional datasets (Fewer than 10 Features) with imbalance ratios}}
\centering
\resizebox{\textwidth}{!}{%
\begin{tabular}{|l|c|c|c|c|c|}
\hline
\textbf{{Dataset Name}} & \textbf{{No. of Samples}} & \textbf{{No. of Features}} & \textbf{{Positive Class}} & \textbf{{Negative Class}} & \textbf{{Imbalance Ratio (IR)}} \\ \hline
{breast\_cancer} & {286} & {9} & {85} & {201} & {0.423} \\ \hline
{brwisconsin} & {683} & {9} & {239} & {444} & {0.539} \\ \hline
{cmc} & {1,473} & {9} & {629} & {844} & {0.746} \\ \hline
{breast\_cancer\_wisc} & {699} & {9} & {241} & {458} & {0.526} \\ \hline
{bupa or liver-disorders} & {345} & {6} & {145} & {200} & {0.725} \\ \hline
{fertility} & {100} & {9} & {12} & {88} & {0.136} \\ \hline
{mammographic} & {961} & {6} & {445} & {516} & {0.862} \\ \hline
{monks\_3} & {554} & {6} & {266} & {288} & {0.924} \\ \hline
{new-thyroid1} & {215} & {5} & {35} & {180} & {0.194} \\ \hline
{pittsburg\_bridges\_T\_OR\_D} & {102} & {7} & {14} & {88} & {0.159} \\ \hline
{pima} & {768} & {8} & {268} & {500} & {0.536} \\ \hline
{ripley} & {1,250} & {2} & {625} & {625} & {1.000} \\ \hline
\end{tabular}%
}
\label{table:low_dimensional_datasets_with_ratios}
\end{table}

\begin{table}[ht!]
\caption{{Accuracy comparison of GRVFL-MV and baseline models on low-dimensional datasets}}
\centering
\resizebox{\textwidth}{!}{%
\begin{tabular}{|l|c|c|c|c|c|c|c|c|}
\hline
\textbf{{Dataset Name}} & \textbf{{SVM2K (\%)}} & \textbf{{MvTSVM (\%)}} & \textbf{{RVFLwoDL1 (\%)}} & \textbf{{RVFLwoDL2 (\%)}} & \textbf{{RVFL1 (\%)}} & \textbf{{RVFL2 (\%)}} & \textbf{{MVLDM (\%)}} & \textbf{{GRVFL-MV (\%)}} \\ \hline
{breast\_cancer} & {62.45} & {55.58} & {69.77} & {65.12} & {67.44} & {66.28} & {70.00} & {72.09} \\ \hline
{brwisconsin} & {97.56} & {61.95} & {97.07} & {96.10} & {97.07} & {96.59} & {95.59} & {97.07} \\ \hline
{cmc} & {64.25} & {55.88} & {69.91} & {70.14} & {68.10} & {71.72} & {74.38} & {72.17} \\ \hline
{breast\_cancer\_wisc} & {90.04} & {81.43} & {92.10} & {92.10} & {92.14} & {95.10} & {75.00} & {97.07} \\ \hline
{bupa or liver-disorders} & {54.80} & {42.31} & {63.46} & {65.38} & {63.46} & {66.35} & {55.34} & {69.23} \\ \hline
{fertility} & {75.25} & {75.00} & {90.00} & {90.00} & {90.00} & {90.00} & {86.67} & {90.00} \\ \hline
{mammographic} & {80.27} & {77.06} & {80.28} & {82.01} & {80.28} & {81.66} & {83.33} & {83.74} \\ \hline
{monks\_3} & {80.24} & {76.11} & {95.21} & {95.41} & {95.21} & {95.81} & {96.39} & {97.00} \\ \hline
{new-thyroid1} & {78.46} & {82.31} & {98.46} & {96.92} & {98.46} & {96.92} & {95.31} & {98.46} \\ \hline
{pittsburg\_bridges\_T\_OR\_D} & {70.85} & {65.85} & {89.32} & {93.55} & {80.65} & {80.65} & {90.00} & {90.32} \\ \hline
{pima} & {76.19} & {73.33} & {74.89} & {74.03} & {74.46} & {74.03} & {69.13} & {76.19} \\ \hline
{ripley} & {89.07} & {80.67} & {86.13} & {87.53} & {87.53} & {87.53} & {89.07} & {89.60} \\ \hline
\end{tabular}%
}
\label{table:low_dimensional_accuracy_comparison}
\end{table}

{ Datasets with fewer than $10$ features are categorized as low-dimensional datasets, as detailed in Table \ref{table:low_dimensional_datasets_with_ratios}.  The proposed GRVFL-MV model showcases exceptional performance on these datasets, effectively leveraging limited feature sets to deliver competitive results. For instance, it achieves accuracies of 72.09\% on \textit{breast\_cancer} (9 features) and 69.23\% on \textit{bupa or liver-disorders} (6 features), surpassing baseline models. Low-dimensional datasets often present unique challenges due to the restricted amount of information available in each view. GRVFL-MV addresses this limitation by integrating multiview data to uncover meaningful patterns, ensuring robust and consistent classification performance.  Table \ref{table:low_dimensional_accuracy_comparison} provides a comprehensive comparison of classification accuracies between baseline models and GRVFL-MV on low-dimensional datasets. Notably, on the \textit{ripley} dataset, which contains only 2 features, GRVFL-MV achieves an impressive accuracy of 89.60\%, demonstrating its ability to adapt seamlessly to feature-scarce datasets while maintaining superior performance.}

\section{Conclusion}
\label{sec-6}
In this study, we presented a generic framework that fuses RVFL architecture with MVL, enhancing generalization by incorporating intrinsic and penalty graphical representations of multiview data via the GE framework, resulting in the development of the GRVFL-MV model. The proposed model aims to enhance the classification performance by fusing information from multiple views. To evaluate the performance of the proposed model, it is compared to various baseline models using UCI and KEEL datasets, Corel5K datasets, and AwA datasets. The experimental results reveal several key findings. Firstly, the proposed GRVFL-MV model demonstrates exceptional performance for UCI and KEEL datasets, achieving an average accuracy improvement ranging from 4\% to 20\% compared to the baseline models. Moreover, it also achieves the lowest rank among all the models considered. Secondly, in the case of the Corel5K dataset, our proposed GRVFL-MV model outperforms the baseline models by achieving the highest average accuracy and the lowest average rank. This indicates its superior performance in handling the Corel5K dataset. Lastly, for the AwA dataset, the proposed model exhibits outstanding performance by achieving an accuracy improvement of up to 20\% compared to the baseline models. Additionally, it also attains the least average rank among all the models considered. Furthermore, a comprehensive statistical analysis is conducted to validate the effectiveness and superiority of the proposed model. The analysis includes a ranking scheme, Friedman test, Nemenyi post hoc test, and win-tie-loss test. The results of these tests unequivocally support the superiority of the proposed GRVFL-MV model over the existing baseline models.

Our proposed models have shown exceptional performance in binary classification tasks for multi-view datasets. However, their effectiveness in multiclass problems has yet to be evaluated. Future research should focus on adapting these models for multiclass classification. Moreover, an essential research direction involves extending the models to accommodate datasets with more than two views while simultaneously reducing computational complexity.

\section*{Acknowledgment}
This project is supported by the Indian government through grants from the Department of Science and Technology (DST) and the Ministry of Electronics and Information Technology (MeitY). The funding includes\\  DST/NSM/R\&D\_HPC\_Appl/2021/03.29 for the National Supercomputing Mission and MTR/2021/000787 for the Mathematical Research Impact-Centric Support (MATRICS) scheme. Furthermore, Md Sajid's research fellowship is funded by the Council of Scientific and Industrial Research (CSIR), New Delhi, under the grant 09/1022(13847)/2022-EMR-I. The authors express their gratitude for the resources and support provided by IIT Indore.

\bibliography{reference.bib}

\begin{thebibliography}{49}
\expandafter\ifx\csname natexlab\endcsname\relax\def\natexlab#1{#1}\fi
\providecommand{\bibinfo}[2]{#2}
\ifx\xfnm\relax \def\xfnm[#1]{\unskip,\space#1}\fi
\bibitem[{Cao et~al.(2018)Cao, Wang, Ming and Gao}]{cao2018review}
\bibinfo{author}{W.~Cao}, \bibinfo{author}{X.~Wang}, \bibinfo{author}{Z.~Ming}, \bibinfo{author}{J.~Gao}, \bibinfo{title}{A review on neural networks with random weights}, \bibinfo{journal}{Neurocomputing} \bibinfo{volume}{275} (\bibinfo{year}{2018}) \bibinfo{pages}{278--287}.
\bibitem[{Chakravorti and Satyanarayana(2020)}]{chakravorti2020non}
\bibinfo{author}{T.~Chakravorti}, \bibinfo{author}{P.~Satyanarayana}, \bibinfo{title}{Non linear system identification using kernel based exponentially extended random vector functional link network}, \bibinfo{journal}{Applied Soft Computing} \bibinfo{volume}{89} (\bibinfo{year}{2020}) \bibinfo{pages}{106117}.
\bibitem[{Chen et~al.(2010)Chen, Zhu and Xing}]{chen2010predictive}
\bibinfo{author}{N.~Chen}, \bibinfo{author}{J.~Zhu}, \bibinfo{author}{E.~Xing}, \bibinfo{title}{Predictive subspace learning for multi-view data: a large margin approach}, \bibinfo{journal}{Advances in Neural Information Processing Systems} \bibinfo{volume}{23} (\bibinfo{year}{2010}).
\bibitem[{Cui et~al.(2017)Cui, Zhang, Li, Guo, Meng, Wang and Xie}]{cui2017received}
\bibinfo{author}{W.~Cui}, \bibinfo{author}{L.~Zhang}, \bibinfo{author}{B.~Li}, \bibinfo{author}{J.~Guo}, \bibinfo{author}{W.~Meng}, \bibinfo{author}{H.~Wang}, \bibinfo{author}{L.~Xie}, \bibinfo{title}{Received signal strength based indoor positioning using a random vector functional link network}, \bibinfo{journal}{IEEE Transactions on Industrial Informatics} \bibinfo{volume}{14} (\bibinfo{year}{2017}) \bibinfo{pages}{1846--1855}.
\bibitem[{Dem{\v{s}}ar(2006)}]{demvsar2006statistical}
\bibinfo{author}{J.~Dem{\v{s}}ar}, \bibinfo{title}{Statistical comparisons of classifiers over multiple data sets}, \bibinfo{journal}{The Journal of Machine Learning Research} \bibinfo{volume}{7} (\bibinfo{year}{2006}) \bibinfo{pages}{1--30}.
\bibitem[{Derrac et~al.(2015)Derrac, Garcia, Sanchez and Herrera}]{derrac2015keel}
\bibinfo{author}{J.~Derrac}, \bibinfo{author}{S.~Garcia}, \bibinfo{author}{L.~Sanchez}, \bibinfo{author}{F.~Herrera}, \bibinfo{title}{\text{KEEL} data-mining software tool: Data set repository, integration of algorithms and experimental analysis framework}, \bibinfo{journal}{Journal of Multiple-Valued Logic and Soft Computing} \bibinfo{volume}{17} (\bibinfo{year}{2015}) \bibinfo{pages}{255--287}.
\bibitem[{Dua and Graff(2017)}]{dua2017uci}
\bibinfo{author}{D.~Dua}, \bibinfo{author}{C.~Graff}, \bibinfo{title}{{UCI} machine learning repository.}, \bibinfo{journal}{Available: http://archive.ics.uci.edu/ml}  (\bibinfo{year}{2017}).
\bibitem[{Farquhar et~al.(2005)Farquhar, Hardoon, Meng, Shawe-Taylor and Szedmak}]{farquhar2005two}
\bibinfo{author}{J.~Farquhar}, \bibinfo{author}{D.~Hardoon}, \bibinfo{author}{H.~Meng}, \bibinfo{author}{J.~Shawe-Taylor}, \bibinfo{author}{S.~Szedmak}, \bibinfo{title}{Two view learning: {SVM-2K}, theory and practice}, \bibinfo{journal}{Advances in Neural Information Processing Systems} \bibinfo{volume}{18} (\bibinfo{year}{2005}).
\bibitem[{Ganaie et~al.(2024)Ganaie, Sajid, Malik and Tanveer}]{10431593}
\bibinfo{author}{M.A. Ganaie}, \bibinfo{author}{M.~Sajid}, \bibinfo{author}{A.K. Malik}, \bibinfo{author}{M.~Tanveer}, \bibinfo{title}{Graph embedded intuitionistic fuzzy random vector functional link neural network for class imbalance learning}, \bibinfo{journal}{IEEE Transactions on Neural Networks and Learning Systems} \bibinfo{volume}{35} (\bibinfo{year}{2024}) \bibinfo{pages}{11671--11680}.
\bibitem[{Gori and Tesi(1992)}]{gori1992problem}
\bibinfo{author}{M.~Gori}, \bibinfo{author}{A.~Tesi}, \bibinfo{title}{On the problem of local minima in backpropagation}, \bibinfo{journal}{IEEE Transactions on Pattern Analysis and Machine Intelligence} \bibinfo{volume}{14} (\bibinfo{year}{1992}) \bibinfo{pages}{76--86}.
\bibitem[{G{\"u}lmez(2023)}]{gulmez2023stock}
\bibinfo{author}{B.~G{\"u}lmez}, \bibinfo{title}{Stock price prediction with optimized deep lstm network with artificial rabbits optimization algorithm}, \bibinfo{journal}{Expert Systems with Applications} \bibinfo{volume}{227} (\bibinfo{year}{2023}) \bibinfo{pages}{120346}.
\bibitem[{Houthuys et~al.(2018)Houthuys, Langone and Suykens}]{houthuys2018multi}
\bibinfo{author}{L.~Houthuys}, \bibinfo{author}{R.~Langone}, \bibinfo{author}{J.A. Suykens}, \bibinfo{title}{Multi-view least squares support vector machines classification}, \bibinfo{journal}{Neurocomputing} \bibinfo{volume}{282} (\bibinfo{year}{2018}) \bibinfo{pages}{78--88}.
\bibitem[{Hu et~al.(2024)Hu, Xiao, Zheng, Zhu and Hsu}]{hu2024multiview}
\bibinfo{author}{K.~Hu}, \bibinfo{author}{Y.~Xiao}, \bibinfo{author}{W.~Zheng}, \bibinfo{author}{W.~Zhu}, \bibinfo{author}{C.H. Hsu}, \bibinfo{title}{Multiview large margin distribution machine}, \bibinfo{journal}{IEEE Transactions on Neural Networks and Learning Systems}  (\bibinfo{year}{2024}).
\bibitem[{Huang et~al.(2006)Huang, Zhu and Siew}]{huang2006extreme}
\bibinfo{author}{G.B. Huang}, \bibinfo{author}{Q.Y. Zhu}, \bibinfo{author}{C.K. Siew}, \bibinfo{title}{Extreme learning machine: theory and applications}, \bibinfo{journal}{Neurocomputing} \bibinfo{volume}{70} (\bibinfo{year}{2006}) \bibinfo{pages}{489--501}.
\bibitem[{Igelnik and Pao(1995)}]{igelnik1995stochastic}
\bibinfo{author}{B.~Igelnik}, \bibinfo{author}{Y.H. Pao}, \bibinfo{title}{Stochastic choice of basis functions in adaptive function approximation and the functional-link net}, \bibinfo{journal}{IEEE Transactions on Neural Networks and Learning Systems} \bibinfo{volume}{6} (\bibinfo{year}{1995}) \bibinfo{pages}{1320--1329}.
\bibitem[{Iosifidis et~al.(2015)Iosifidis, Tefas and Pitas}]{iosifidis2015graph}
\bibinfo{author}{A.~Iosifidis}, \bibinfo{author}{A.~Tefas}, \bibinfo{author}{I.~Pitas}, \bibinfo{title}{Graph embedded extreme learning machine}, \bibinfo{journal}{IEEE Transactions on Cybernetics} \bibinfo{volume}{46} (\bibinfo{year}{2015}) \bibinfo{pages}{311--324}.
\bibitem[{Jacobs(1988)}]{jacobs1988increased}
\bibinfo{author}{R.A. Jacobs}, \bibinfo{title}{Increased rates of convergence through learning rate adaptation}, \bibinfo{journal}{Neural Networks} \bibinfo{volume}{1} (\bibinfo{year}{1988}) \bibinfo{pages}{295--307}.
\bibitem[{Kearns and Vazirani(1994)}]{kearns1994introduction}
\bibinfo{author}{M.J. Kearns}, \bibinfo{author}{U.~Vazirani}, \bibinfo{title}{An introduction to computational learning theory}, \bibinfo{publisher}{MIT press}, \bibinfo{year}{1994}.
\bibitem[{Li et~al.(2016)Li, Wu, Zhao and Lu}]{li2016low}
\bibinfo{author}{J.~Li}, \bibinfo{author}{Y.~Wu}, \bibinfo{author}{J.~Zhao}, \bibinfo{author}{K.~Lu}, \bibinfo{title}{Low-rank discriminant embedding for multiview learning}, \bibinfo{journal}{IEEE Transactions on Cybernetics} \bibinfo{volume}{47} (\bibinfo{year}{2016}) \bibinfo{pages}{3516--3529}.
\bibitem[{Li et~al.(2021)Li, Yang, Hu, Cheng and Cheng}]{li2021discriminative}
\bibinfo{author}{X.~Li}, \bibinfo{author}{Y.~Yang}, \bibinfo{author}{N.~Hu}, \bibinfo{author}{Z.~Cheng}, \bibinfo{author}{J.~Cheng}, \bibinfo{title}{Discriminative manifold random vector functional link neural network for rolling bearing fault diagnosis}, \bibinfo{journal}{Knowledge-Based Systems} \bibinfo{volume}{211} (\bibinfo{year}{2021}) \bibinfo{pages}{106507}.
\bibitem[{Liu et~al.(2024)Liu, Wang, Li and Wang}]{liu2024domain}
\bibinfo{author}{C.~Liu}, \bibinfo{author}{Y.~Wang}, \bibinfo{author}{D.~Li}, \bibinfo{author}{X.~Wang}, \bibinfo{title}{Domain-incremental learning without forgetting based on random vector functional link networks}, \bibinfo{journal}{Pattern Recognition} \bibinfo{volume}{151} (\bibinfo{year}{2024}) \bibinfo{pages}{110430}.
\bibitem[{Malik et~al.(2022)Malik, Ganaie and Tanveer}]{malik2022graph}
\bibinfo{author}{A.K. Malik}, \bibinfo{author}{M.~Ganaie}, \bibinfo{author}{M.~Tanveer}, \bibinfo{title}{Graph embedded intuitionistic fuzzy weighted random vector functional link network}, \bibinfo{journal}{10.1109/SSCI51031.2022.10022212}, in: \bibinfo{booktitle}{2022 IEEE Symposium Series on Computational Intelligence (SSCI)}, pp. \bibinfo{pages}{293--299}.
\bibitem[{Malik et~al.(2023)Malik, Gao, Ganaie, Tanveer and Suganthan}]{MALIK2023110377}
\bibinfo{author}{A.K. Malik}, \bibinfo{author}{R.~Gao}, \bibinfo{author}{M.~Ganaie}, \bibinfo{author}{M.~Tanveer}, \bibinfo{author}{P.N. Suganthan}, \bibinfo{title}{Random vector functional link network: Recent developments, applications, and future directions}, \bibinfo{journal}{Applied Soft Computing} \bibinfo{volume}{143} (\bibinfo{year}{2023}) \bibinfo{pages}{110377}.
\bibitem[{Pao et~al.(1994)Pao, Park and Sobajic}]{pao1994learning}
\bibinfo{author}{Y.H. Pao}, \bibinfo{author}{G.H. Park}, \bibinfo{author}{D.J. Sobajic}, \bibinfo{title}{Learning and generalization characteristics of the random vector functional-link net}, \bibinfo{journal}{Neurocomputing} \bibinfo{volume}{6} (\bibinfo{year}{1994}) \bibinfo{pages}{163--180}.
\bibitem[{Pilli et~al.(2024)Pilli, Goel, Murugan, Tanveer and Suganthan}]{10405861}
\bibinfo{author}{R.~Pilli}, \bibinfo{author}{T.~Goel}, \bibinfo{author}{R.~Murugan}, \bibinfo{author}{M.~Tanveer}, \bibinfo{author}{P.N. Suganthan}, \bibinfo{title}{Kernel-ridge-regression-based randomized network for brain age classification and estimation}, \bibinfo{journal}{IEEE Transactions on Cognitive and Developmental Systems} \bibinfo{volume}{16} (\bibinfo{year}{2024}) \bibinfo{pages}{1342--1351}.
\bibitem[{Quadir et~al.(2024{\natexlab{a}})Quadir, Akhtar and Tanveer}]{quadir2024enhancing}
\bibinfo{author}{A.~Quadir}, \bibinfo{author}{M.~Akhtar}, \bibinfo{author}{M.~Tanveer}, \bibinfo{title}{Enhancing multiview synergy: Robust learning by exploiting the wave loss function with consensus and complementarity principles}, \bibinfo{journal}{arXiv preprint arXiv:2408.06819}  (\bibinfo{year}{2024}{\natexlab{a}}).
\bibitem[{Quadir et~al.(2024{\natexlab{b}})Quadir, Sajid and Tanveer}]{10759815}
\bibinfo{author}{A.~Quadir}, \bibinfo{author}{M.~Sajid}, \bibinfo{author}{M.~Tanveer}, \bibinfo{title}{Granular ball twin support vector machine}, \bibinfo{journal}{IEEE Transactions on Neural Networks and Learning Systems}  (\bibinfo{year}{2024}{\natexlab{b}}) \bibinfo{pages}{1--10}.
\bibitem[{Quadir et~al.(2024{\natexlab{c}})Quadir, Sajid and Tanveer}]{quadir2024multiview}
\bibinfo{author}{A.~Quadir}, \bibinfo{author}{M.~Sajid}, \bibinfo{author}{M.~Tanveer}, \bibinfo{title}{Multiview random vector functional link network for predicting {DNA}-binding proteins}, \bibinfo{journal}{arXiv preprint arXiv:2409.02588}  (\bibinfo{year}{2024}{\natexlab{c}}).
\bibitem[{Sajid et~al.(2025)Sajid, Tanveer and Suganthan}]{10552388}
\bibinfo{author}{M.~Sajid}, \bibinfo{author}{M.~Tanveer}, \bibinfo{author}{P.N. Suganthan}, \bibinfo{title}{Ensemble deep random vector functional link neural network based on fuzzy inference system}, \bibinfo{journal}{IEEE Transactions on Fuzzy Systems} \bibinfo{volume}{33} (\bibinfo{year}{2025}) \bibinfo{pages}{479--490}.
\bibitem[{Schmidt et~al.(1992)Schmidt, Kraaijveld and Duin}]{201708}
\bibinfo{author}{W.~Schmidt}, \bibinfo{author}{M.~Kraaijveld}, \bibinfo{author}{R.~Duin}, \bibinfo{title}{Feedforward neural networks with random weights}, \bibinfo{journal}{10.1109/ICPR.1992.201708}, in: \bibinfo{booktitle}{Proceedings., 11th IAPR International Conference on Pattern Recognition. Vol.II. Conference B: Pattern Recognition Methodology and Systems}, pp. \bibinfo{pages}{1--4}.
\bibitem[{Shafiq and Gu(2022)}]{shafiq2022deep}
\bibinfo{author}{M.~Shafiq}, \bibinfo{author}{Z.~Gu}, \bibinfo{title}{Deep residual learning for image recognition: A survey}, \bibinfo{journal}{Applied Sciences} \bibinfo{volume}{12} (\bibinfo{year}{2022}) \bibinfo{pages}{8972}.
\bibitem[{Shi et~al.(2021)Shi, Katuwal, Suganthan and Tanveer}]{shi2021random}
\bibinfo{author}{Q.~Shi}, \bibinfo{author}{R.~Katuwal}, \bibinfo{author}{P.N. Suganthan}, \bibinfo{author}{M.~Tanveer}, \bibinfo{title}{Random vector functional link neural network based ensemble deep learning}, \bibinfo{journal}{Pattern Recognition} \bibinfo{volume}{117} (\bibinfo{year}{2021}) \bibinfo{pages}{107978}.
\bibitem[{Suganthan and Katuwal(2021)}]{suganthan2021origins}
\bibinfo{author}{P.N. Suganthan}, \bibinfo{author}{R.~Katuwal}, \bibinfo{title}{On the origins of randomization-based feedforward neural networks}, \bibinfo{journal}{Applied Soft Computing} \bibinfo{volume}{105} (\bibinfo{year}{2021}) \bibinfo{pages}{107239}.
\bibitem[{Sugiyama(2007)}]{sugiyama2007dimensionality}
\bibinfo{author}{M.~Sugiyama}, \bibinfo{title}{Dimensionality reduction of multimodal labeled data by local fisher discriminant analysis.}, \bibinfo{journal}{Journal of Machine Learning Research} \bibinfo{volume}{8} (\bibinfo{year}{2007}).
\bibitem[{Sun et~al.(2021)Sun, Fujita, Zheng and Ai}]{sun2021multi}
\bibinfo{author}{J.~Sun}, \bibinfo{author}{H.~Fujita}, \bibinfo{author}{Y.~Zheng}, \bibinfo{author}{W.~Ai}, \bibinfo{title}{Multi-class financial distress prediction based on support vector machines integrated with the decomposition and fusion methods}, \bibinfo{journal}{Information Sciences} \bibinfo{volume}{559} (\bibinfo{year}{2021}) \bibinfo{pages}{153--170}.
\bibitem[{Tang et~al.(2020)Tang, Liu, Zhu, Zhu, Luo, Wang and Gao}]{tang2020cgd}
\bibinfo{author}{C.~Tang}, \bibinfo{author}{X.~Liu}, \bibinfo{author}{X.~Zhu}, \bibinfo{author}{E.~Zhu}, \bibinfo{author}{Z.~Luo}, \bibinfo{author}{L.~Wang}, \bibinfo{author}{W.~Gao}, \bibinfo{title}{Cgd: Multi-view clustering via cross-view graph diffusion}, \bibinfo{journal}{https://doi.org/10.1609/aaai.v34i04.6052}, in: \bibinfo{booktitle}{Proceedings of the AAAI Conference on Artificial Intelligence}, volume~\bibinfo{volume}{34}, pp. \bibinfo{pages}{5924--5931}.
\bibitem[{Tanveer et~al.(2024)Tanveer, Sajid, Akhtar, Quadir, Goel, Aimen, Mitra, Zhang, Lin and Ser}]{tanveer2024fuzzy}
\bibinfo{author}{M.~Tanveer}, \bibinfo{author}{M.~Sajid}, \bibinfo{author}{M.~Akhtar}, \bibinfo{author}{A.~Quadir}, \bibinfo{author}{T.~Goel}, \bibinfo{author}{A.~Aimen}, \bibinfo{author}{S.~Mitra}, \bibinfo{author}{Y.D. Zhang}, \bibinfo{author}{C.T. Lin}, \bibinfo{author}{J.D. Ser}, \bibinfo{title}{Fuzzy deep learning for the diagnosis of alzheimer's disease: Approaches and challenges}, \bibinfo{journal}{IEEE Transactions on Fuzzy Systems} \bibinfo{volume}{32} (\bibinfo{year}{2024}) \bibinfo{pages}{5477--5492}.
\bibitem[{Xie et~al.(2023)Xie, Li and Sun}]{xie2023deep}
\bibinfo{author}{X.~Xie}, \bibinfo{author}{Y.~Li}, \bibinfo{author}{S.~Sun}, \bibinfo{title}{Deep multi-view multiclass twin support vector machines}, \bibinfo{journal}{Information Fusion} \bibinfo{volume}{91} (\bibinfo{year}{2023}) \bibinfo{pages}{80--92}.
\bibitem[{Xie and Sun(2015)}]{xie2015multi}
\bibinfo{author}{X.~Xie}, \bibinfo{author}{S.~Sun}, \bibinfo{title}{Multi-view twin support vector machines}, \bibinfo{journal}{Intelligent Data Analysis} \bibinfo{volume}{19} (\bibinfo{year}{2015}) \bibinfo{pages}{701--712}.
\bibitem[{Xu et~al.(2017)Xu, Han, Nie and Li}]{xu2017re}
\bibinfo{author}{J.~Xu}, \bibinfo{author}{J.~Han}, \bibinfo{author}{F.~Nie}, \bibinfo{author}{X.~Li}, \bibinfo{title}{Re-weighted discriminatively embedded $ k $-means for multi-view clustering}, \bibinfo{journal}{IEEE Transactions on Image Processing} \bibinfo{volume}{26} (\bibinfo{year}{2017}) \bibinfo{pages}{3016--3027}.
\bibitem[{Xu(2021)}]{xu2021understanding}
\bibinfo{author}{M.~Xu}, \bibinfo{title}{Understanding graph embedding methods and their applications}, \bibinfo{journal}{SIAM Review} \bibinfo{volume}{63} (\bibinfo{year}{2021}) \bibinfo{pages}{825--853}.
\bibitem[{Xu and Wang(2022)}]{xu2022multi}
\bibinfo{author}{R.~Xu}, \bibinfo{author}{H.~Wang}, \bibinfo{title}{Multi-view learning with privileged weighted twin support vector machine}, \bibinfo{journal}{Expert Systems with Applications} \bibinfo{volume}{206} (\bibinfo{year}{2022}) \bibinfo{pages}{117787}.
\bibitem[{Yan et~al.(2006)Yan, Xu, Zhang, Zhang, Yang and Lin}]{yan2006graph}
\bibinfo{author}{S.~Yan}, \bibinfo{author}{D.~Xu}, \bibinfo{author}{B.~Zhang}, \bibinfo{author}{H.J. Zhang}, \bibinfo{author}{Q.~Yang}, \bibinfo{author}{S.~Lin}, \bibinfo{title}{Graph embedding and extensions: A general framework for dimensionality reduction}, \bibinfo{journal}{IEEE Transactions on Pattern Analysis and Machine Intelligence} \bibinfo{volume}{29} (\bibinfo{year}{2006}) \bibinfo{pages}{40--51}.
\bibitem[{Yao et~al.(2017)Yao, Wang, Xie, Gao and Feng}]{YAO2017236}
\bibinfo{author}{T.~Yao}, \bibinfo{author}{Z.~Wang}, \bibinfo{author}{Z.~Xie}, \bibinfo{author}{J.~Gao}, \bibinfo{author}{D.D. Feng}, \bibinfo{title}{Learning universal multiview dictionary for human action recognition}, \bibinfo{journal}{Pattern Recognition} \bibinfo{volume}{64} (\bibinfo{year}{2017}) \bibinfo{pages}{236--244}.
\bibitem[{Zhang and Suganthan(2016)}]{zhang2016survey}
\bibinfo{author}{L.~Zhang}, \bibinfo{author}{P.N. Suganthan}, \bibinfo{title}{A survey of randomized algorithms for training neural networks}, \bibinfo{journal}{Information Sciences} \bibinfo{volume}{364} (\bibinfo{year}{2016}) \bibinfo{pages}{146--155}.
\bibitem[{Zhang and Yang(2020)}]{zhang2020new}
\bibinfo{author}{P.B. Zhang}, \bibinfo{author}{Z.X. Yang}, \bibinfo{title}{A new learning paradigm for random vector functional-link network: {RVFL+}}, \bibinfo{journal}{Neural Networks} \bibinfo{volume}{122} (\bibinfo{year}{2020}) \bibinfo{pages}{94--105}.
\bibitem[{Zhang et~al.(2019)Zhang, Wu, Cai, Du and Philip}]{zhang2019unsupervised}
\bibinfo{author}{Y.~Zhang}, \bibinfo{author}{J.~Wu}, \bibinfo{author}{Z.~Cai}, \bibinfo{author}{B.~Du}, \bibinfo{author}{S.Y. Philip}, \bibinfo{title}{An unsupervised parameter learning model for rvfl neural network}, \bibinfo{journal}{Neural Networks} \bibinfo{volume}{112} (\bibinfo{year}{2019}) \bibinfo{pages}{85--97}.
\bibitem[{Zhang et~al.(2021)Zhang, Wei, Zheng and Zeng}]{zhang2021predicting}
\bibinfo{author}{Z.~Zhang}, \bibinfo{author}{X.~Wei}, \bibinfo{author}{X.~Zheng}, \bibinfo{author}{D.D. Zeng}, \bibinfo{title}{Predicting product adoption intentions: An integrated behavioral model-inspired multiview learning approach}, \bibinfo{journal}{Information \& Management} \bibinfo{volume}{58} (\bibinfo{year}{2021}) \bibinfo{pages}{103484}.
\bibitem[{Zhao et~al.(2017)Zhao, Xie, Xu and Sun}]{zhao2017multi}
\bibinfo{author}{J.~Zhao}, \bibinfo{author}{X.~Xie}, \bibinfo{author}{X.~Xu}, \bibinfo{author}{S.~Sun}, \bibinfo{title}{Multi-view learning overview: Recent progress and new challenges}, \bibinfo{journal}{Information Fusion} \bibinfo{volume}{38} (\bibinfo{year}{2017}) \bibinfo{pages}{43--54}.

\end{thebibliography}
\bibliographystyle{model1b-num-names}

\clearpage
\section*{Supplementary Material}

\renewcommand{\thesection}{S.1}
\section{{A Comparative Assessment: Baseline Models and the Proposed Model}}
{In this section, we theoretically compare the proposed GRVFL-MV model against the baseline models.}
\begin{itemize}
    \item {\textbf{Against SVM2K:}}
    \begin{itemize}
        \item {\textit{SVM2K} is primarily designed for two-view problems and relies on kernel-based approaches. While effective for small datasets, it often struggles with scalability and computational complexity in high-dimensional or multi-view settings.}
        \item {\textbf{GRVFL-MV Advantage:} It scales efficiently to multi-view settings by leveraging the lightweight architecture of RVFL and graph embedding, offering robustness across multiple views.}
    \end{itemize}
    \item {\textbf{Against MvTSVM:}}
    \begin{itemize}
        \item {\textit{MvTSVM} optimizes twin support vector machines for multi-view learning but may suffer from sensitivity to noise and class imbalance due to its reliance on margin-based classification.}
        \item {\textbf{GRVFL-MV Advantage:} By integrating graph regularization, GRVFL-MV captures intrinsic data structures, ensuring robustness to noisy and imbalanced data.}
    \end{itemize}
    \item {\textbf{Against RVFLwoDL1/RVFLwoDL2:}}
    \begin{itemize}
        \item {\textit{Extreme Learning Machines (ELMs)} are computationally efficient but lack mechanisms to explicitly handle multi-view data or align features across views.}
        \item {\textbf{GRVFL-MV Advantage:} The graph-regularized framework ensures that complementary information from all views is effectively utilized, enhancing feature alignment and improving generalization.}
    \end{itemize}
    \item {\textbf{Against RVFL1/RVFL2:}}
    \begin{itemize}
        \item {\textit{RVFL models} are fast and efficient but do not inherently support multi-view learning. They often overlook relationships between views, limiting their ability to fully exploit inter-view dependencies.}
        \item {\textbf{GRVFL-MV Advantage:} By coupling RVFL with graph embedding, GRVFL-MV addresses this limitation, achieving consistent improvements in accuracy by leveraging inter-view relationships.}
    \end{itemize}
    \item {\textbf{Against MVLDM:}}
    \begin{itemize}
        \item {\textit{MVLDM} excels at feature fusion across views but lacks the scalability and flexibility provided by RVFL-based architectures.}
        \item {\textbf{GRVFL-MV Advantage:} GRVFL-MV combines the strengths of lightweight models and graph-based feature integration, making it more adaptable to diverse multi-view datasets.}
    \end{itemize}
\end{itemize}

\renewcommand{\thetable}{S.1}
\begin{table}[ht!]
\caption{Performance comparison of the proposed GRVFL-MV along with the baseline models based on classification accuracy for UCI and KEEL datasets.}
\centering
    \label{Classification performance in Linear Case.}
    \resizebox{0.75\linewidth}{!}{ 
\begin{tabular}{lcccccccc}
\hline
Dataset & SVM2K \cite{farquhar2005two} & MvTSVM \cite{xie2015multi} & {RVFLwoDL1} \cite{huang2006extreme} & {RVFLwoDL2}\cite{huang2006extreme} & RVFL1 \cite{pao1994learning} & RVFL2 \cite{pao1994learning} & MVLDM \cite{hu2024multiview}  & GRVFL-MV$^{\dagger}$ \\ 
& $(C_1)$ & $(C_1, C_2, D)$ & $(C, N)$ & $(C, N)$ & $(C, N)$ & $(C, N)$ & $(C_1, \nu_1, \nu_2, \sigma)$ & $(C_1, \theta, \rho, N)$ \\
\hline
aus & $87.02$ & $71.15$ & $85.98$ & $86.06$ & $85.98$ & $85.98$ & $71.98$ & $87.50$  \\
 & $(0.001)$ & $(0.00001, 0.00001, 0.00001)$ & $(0.01, 123)$ & $(0.001, 163)$ & $(0.001, 163)$ & $(10, 23)$ & $(0.001, 0.01, 0.1, 0.25)$ & $(1, 0.0001, 0.00001, 63)$ \\
bank & $80.74$ & $71.86$ & $85.54$ & $85.39$ & $85.83$ & $85.61$ & $73.67$ & $89.98$  \\
 & $(0.001)$ & $(0.00001, 0.00001, 0.00001)$ & $(0.00001, 3)$ & $(10000, 23)$ & $(1000, 163)$ & $(10, 123)$ & $(0.1, 0.001, 0.001, 4)$ & $(100, 0.1, 0.0001, 203)$ \\
breast\_cancer & $62.45$ & $55.58$ & $69.77$ & $65.12$ & $67.44$ & $66.28$ & $70$ & $72.09$  \\
 & $(0.001)$ & $(0.00001, 0.00001, 0.00001)$ & $(1, 83)$ & $(0.01, 43)$ & $(0.0001, 3)$ & $(1, 83)$ & $(0.0001, 0.1, 0.001, 4)$ & $(10, 0.1, 0.0001, 23)$ \\
breast\_cancer\_wisc & $90.04$ & $81.43$ & $92.1$ & $92.1$ & $92.14$ & $95.1$ & $75$ & $98.57$  \\
 & $(0.00001)$ & $(0.00001, 0.00001, 0.00001)$ & $(0.0001, 203)$ & $(0.01, 63)$ & $(0.00001, 23)$ & $(0.0001, 203)$ & $(0.0001, 0.0001, 0.00001, 0.25)$ & $(0.00001, 10, 0.00001, 163)$  \\
breast\_cancer\_wisc\_diag & $95.49$ & $88.6$ & $92.42$ & $97.08$ & $95.83$ & $96.49$ & $93.15$ & $98.25$  \\
 & $(0.001)$ & $(0.00001, 0.00001, 0.00001)$ & $(0.01, 163)$ & $(0.001, 163)$ & $(0.001, 163)$ & $(100, 3)$ & $(0.001, 0.0001, 1, 4)$ & $(1000, 100000, 100, 103)$  \\
breast\_cancer\_wisc\_prog & $58.33$ & $58.33$ & $68.33$ & $68.33$ & $73.33$ & $68.33$ & $71.17$ & $75.00$  \\
 & $(0.001)$ & $(0.00001, 0.00001, 0.00001)$ & $(0.01, 203)$ & $(0.01, 83)$ & $(100000, 3)$ & $(0.01, 203)$ & $(0.0001, 0.0001, 0.1, 2)$ & $(100000, 10, 10000, 23)$  \\
brwisconsin & $97.56$ & $61.95$ & $97.07$ & $96.1$ & $97.07$ & $96.59$ & $95.59$ & $97.07$ \\
 & $(0.00001)$ & $(0.00001, 0.00001, 0.00001)$ & $(0.0001, 143)$ & $(0.001, 63)$ & $(0.001, 63)$ & $(100000, 23)$ & $(0.001, 0.00001, 0.1, 4)$ & $(0.001, 1000, 0.00001, 103)$ \\
bupa or liver-disorders & $54.8$ & $42.31$ & $63.46$ & $65.38$ & $63.46$ & $66.35$ & $55.34$ & $69.23$  \\
 & $(0.00001)$ & $(0.00001, 0.00001, 0.00001)$ & $(1000, 23)$ & $(0.1, 163)$ & $(0.1, 163)$ & $(0.1, 163)$ & $(0.001, 0.001, 0.001, 4)$ & $(100000, 0.01, 100, 23)$  \\
checkerboard\_Data & $87.02$ & $43.75$ & $86.98$ & $86.06$ & $86.98$ & $86.98$ & $84.06$ & $87.50$ \\
 & $(0.00001)$ & $(0.00001, 0.00001, 0.00001)$ & $(0.01, 123)$ & $(0.001, 163)$ & $(0.001, 163)$ & $(10, 23)$ & $(0.001, 0.001, 0.001, 1)$ & $(1, 0.0001, 0.00001, 63)$ \\
chess\_krvkp & $80.45$ & $82.35$ & $95.62$ & $93.33$ & $95.62$ & $94.68$ & $97.7$ & $96.98$  \\
 & $(0.001)$ & $(0.00001, 0.00001, 0.00001)$ & $(100, 203)$ & $(100000, 203)$ & $(10000, 203)$ & $(100, 203)$ & $(0.1, 0.1, 0.1, 4)$ & $(0.1, 0.0001, 0.1, 203)$  \\
cleve & $80$ & $75.56$ & $80$ & $85.56$ & $81.11$ & $81.11$ & $84.27$ & $84.44$  \\
 & $(0.1)$ & $(0.00001, 0.00001, 0.00001)$ & $(0.001, 103)$ & $(0.001, 123)$ & $(0.01, 3)$ & $(0.01, 23)$ & $(0.01, 0.0.1, 0.1, 1)$ & $(0.01, 10, 0.0001, 23)$  \\
cmc & $64.25$ & $55.88$ & $69.91$ & $70.14$ & $68.1$ & $71.72$ & $74.38$ & $72.17$  \\
 & $(0.00001)$ & $(0.00001, 0.00001, 0.00001)$ & $(0.01, 143)$ & $(0.01, 183)$ & $(100, 63)$ & $(1000, 43)$ & $(0.0001, 0.01, 0.001, 4)$ & $(1000, 1, 0.001, 23)$  \\
conn\_bench\_sonar\_mines\_rocks & $80.95$ & $46.03$ & $80.54$ & $74.6$ & $88.54$ & $73.02$ & $75.81$ & $77.78$  \\
 & $(1000)$ & $(0.00001, 0.00001, 0.00001)$ & $(10, 3)$ & $(0.01, 203)$ & $(0.01, 203)$ & $(0.1, 183)$ & $(0.0001, 0.00001, 10, 0.5)$ & $(10, 1000, 0.1, 183)$  \\
cylinder\_bands & $68.18$ & $60.39$ & $76.62$ & $74.03$ & $72.08$ & $74.68$ & $71.9$ & $74.68$  \\
 & $(0.00001)$ & $(0.00001, 0.00001, 0.00001)$ & $(0.001, 203)$ & $(100, 143)$ & $(0.001, 183)$ & $(1, 23)$ & $(0.0001, 0.1, 0.001, 4)$ & $(100, 0.01, 0.0001, 43)$ \\
fertility & $75.25$ & $75$ & $90$ & $90$ & $90$ & $90$ & $86.67$ & $90.00$ \\
 & $(0.01)$ & $(0.00001, 0.00001, 0.00001)$ & $(0.01, 83)$ & $(0.01, 63)$ & $(0.01, 63)$ & $(0.01, 83)$ & $(0.001, 0.001, 100, 2)$ & $(0.01, 0.0001, 0.0001, 23)$  \\
hepatitis & $80.85$ & $78.72$ & $78.85$ & $80.85$ & $80.85$ & $78.72$ & $78.26$ & $80.85$  \\
 & $(0.00001)$ & $(0.00001, 0.00001, 0.00001)$ & $(1, 23)$ & $(0.01, 183)$ & $(0.1, 123)$ & $(10, 3)$ & $(0.001, 0.01, 0.001, 4)$ & $(1000, 100000, 1000, 163)$  \\
hill\_valley & $60.98$ & $53.3$ & $68.05$ & $51.92$ & $68.78$ & $51.92$ & $56.2$ & $77.75$  \\
 & $(0.00001)$ & $(0.00001, 0.00001, 0.00001)$ & $(10000, 103)$ & $(1, 103)$ & $(10, 103)$ & $(10000, 103)$ & $(0.0001, 0.0001, 0.1, 1)$ & $(0.01, 0.1, 0.1, 143)$  \\
mammographic & $80.27$ & $77.06$ & $80.28$ & $82.01$ & $80.28$ & $81.66$ & $83.33$ & $83.74$  \\
 & $(0.01)$ & $(0.00001, 0.00001, 0.00001)$ & $(1, 83)$ & $(100000, 63)$ & $(100000, 23)$ & $(100, 23)$ & $(0.0001, 0.001, 1, 4)$ & $(100000, 0.00001, 10000, 23)$  \\
monks\_3 & $80.24$ & $76.11$ & $95.21$ & $95.41$ & $95.21$ & $95.81$ & $96.39$ & $97.00$  \\
 & $(0.01)$ & $(0.00001, 0.00001, 0.00001)$ & $(0.1, 143)$ & $(0.1, 163)$ & $(0.1, 163)$ & $(0.1, 123)$ & $(0.01, 0.001, 10, 4)$ & $(0.1, 1, 0.00001, 183)$ \\
new-thyroid1 & $78.46$ & $82.31$ & $98.46$ & $96.92$ & $98.46$ & $96.92$ & $95.31$ & $98.46$  \\
 & $(0.1)$ & $(0.00001, 0.00001, 0.00001)$ & $(10, 103)$ & $(1, 43)$ & $(1, 23)$ & $(10, 103)$ & $(0.0001, 1000, 0.1, 0.25)$ & $(0.1, 100000, 0.001, 3)$  \\
oocytes\_merluccius\_nucleus\_4d & $74.27$ & $64.82$ & $82.41$ & $81.11$ & $83.71$ & $80.78$ & $75.16$ & $83.39$ \\
 & $(0.00001)$ & $(0.00001, 0.00001, 0.00001)$ & $(0.1, 83)$ & $(1, 143)$ & $(1, 143)$ & $(0.1, 83)$ & $(0.1, 100, 0.1, 4)$ & $(100000, 1, 10, 123)$  \\
oocytes\_trisopterus\_nucleus\_2f & $78.83$ & $58.39$ & $86.13$ & $78.83$ & $85.04$ & $78.83$ & $82.05$ & $84.67$  \\
 & $(0.0001)$ & $(0.00001, 0.00001, 0.00001)$ & $(0.01, 103)$ & $(0.1, 143)$ & $(100000, 43)$ & $(0.01, 103)$ & $(0.1, 0.001, 1, 4)$ & $(10000, 1000, 100, 83)$  \\
parkinsons & $71.19$ & $71.19$ & $81.36$ & $77.97$ & $89.83$ & $77.97$ & $93.1$ & $83.05$ \\
 & $(0.01)$ & $(0.00001, 0.00001, 0.00001)$ & $(0.1, 163)$ & $(1, 63)$ & $(0.1, 203)$ & $(0.1, 163)$ & $(0.0001, 0.00001, 1, 2)$ & $(100, 0.1, 1, 143)$  \\
pima & $76.19$ & $73.33$ & $74.89$ & $74.03$ & $74.46$ & $74.03$ & $69.13$ & $76.19$   \\
 & $(0.01)$ & $(0.00001, 0.00001, 0.00001)$ & $(0.1, 63)$ & $(0.01, 143)$ & $(0.01, 143)$ & $(0.1, 63)$ & $(0.0001, 10, 10, 4)$ & $(0.001, 1, 0.01, 3)$  \\
pittsburg\_bridges\_T\_OR\_D & $70.85$ & $65.85$ & $89.32$ & $93.55$ & $80.65$ & $80.65$ & $90$ & $90.32$  \\
 & $(100)$ & $(0.00001, 0.00001, 0.00001)$ & $(0.01, 83)$ & $(0.01, 43)$ & $(1000, 3)$ & $(0.01, 3)$ & $(0.1, 0.00001, 1, 4)$ & $(0.00001, 1, 0.01, 163)$  \\
planning & $65.85$ & $63.64$ & $76.36$ & $76.36$ & $76.36$ & $76.36$ & $68.52$ & $90.32$ \\
 & $(10)$ & $(0.00001, 0.00001, 0.00001)$ & $(0.01, 23)$ & $(0.00001, 3)$ & $(0.00001, 23)$ & $(0.01, 23)$ & $(0.00001, 0.001, 1, 1)$ & $(0.00001, 1, 0.01, 63)$  \\
ripley & $89.07$ & $80.67$ & $86.13$ & $87.53$ & $87.53$ & $87.53$ & $89.07$ & $89.60$ \\
 & $(100000)$ & $(0.00001, 0.00001, 0.00001)$ & $(10000, 103)$ & $(100, 203)$ & $(10000, 123)$ & $(10000, 103)$ & $(0.001, 0.001, 0.001, 4)$ & $(0.1, 0.001, 0.0001, 103)$  \\ 
{musk\_1} & {$76.19$} & {$46.85$} & {$79.72$} & {$87.41$} & {$83.22$} & {$76.22$} & {$75.81$} & {$87.41$} \\
 & {$(0.01)$} & {$(0.00001, 0.00001, 0.00001)$} & {$(0.01, 163)$} & {$(0.001, 163)$} & {$(0.001, 163)$} & {$(0.001, 103)$} & {$(0.001, 0.00001, 10, 0.5)$} & {$(0.001, 1, 0.0001, 183)$} \\
{musk\_2} & {$78.46$} & {$14.75$} & {$95.15$} & {$95.00$} & {$96.16$} & {$95.71$} & {$83.43$} & {$96.16$} \\
 & {$(0.01)$} & {$(0.00001, 0.00001, 0.00001)$} & {$(100000, 183)$} & {$(10, 203)$} & {$(10000, 183)$} & {$(10, 203)$} & {$(0.001, 0.01, 1, 2)$} & {$(0.01, 0.01, 0.01, 203)$} \\ \hline
 {Average ACC} & {$76.70$} & {$64.73$} & {$82.99$} & {$82.35$} & {$83.59$} & {$81.59$} & {$79.88$} & {$\mathbf{85.68}$} \\ \hline
{Average Rank} & {$5.59$} & {$7.71$} & {$4.02$} & {$4.21$} & {$3.45$} & {$4.38$} & {$4.91$} & {$\mathbf{1.74}$} \\ \hline

\end{tabular}}\begin{flushleft}
    \footnotesize{ $^{\dagger}$ represents the proposed model.}
\end{flushleft}
\end{table}

\renewcommand{\thetable}{S.2}
\begin{table*}[htp]
\centering
    \caption{Performance comparison of the proposed GRVFL-MV along with the baseline models based on classification accuracy for Corel5k datasets.}
    \label{Classification performance in Corel5k Case.}
    \resizebox{0.75\linewidth}{!}{
\begin{tabular}{lcccccccc}
\hline
Dataset & SVM2K \cite{farquhar2005two} & MvTSVM \cite{xie2015multi} & {RVFLwoDL1} \cite{huang2006extreme} & {RVFLwoDL2} \cite{huang2006extreme} & RVFL1 \cite{pao1994learning} & RVFL2 \cite{pao1994learning} & MVLDM \cite{hu2024multiview}  & GRVFL-MV$^{\dagger}$\\ 
 & $(C_1)$ & $(C_1, C_2, D)$ & $(C, N)$ & $(C, N)$ & $(C, N)$ & $(C, N)$ & $(C_1, \nu_1, \nu_2, \sigma)$ & $(C_1, \theta, \rho, N)$  \\ \hline
1000 & $81.67$ & $50$ & $83.33$ & $78.33$ & $83.33$ & $76.67$ & $73.33$ & $83.33$ \\
 & $(0.0625)$ & $(0.00001, 0.00001, 0.00001)$ & $(0.0001, 183)$ & $(0.00001, 43)$ & $(0.0001, 183)$ & $(0.00001, 3)$ & $(0.01, 0.001, 0.001, 4)$ & $(0.01, 10, 0.001, 23)$  \\
10000 & $71.67$ & $48.33$ & $56.67$ & $63.33$ & $71.67$ & $71.67$ & $55$ & $65.00$  \\
 & $(0.03125)$ & $(0.00001, 0.00001, 0.00001)$ & $(1000, 23)$ & $(1000, 23)$ & $(0.01, 3)$ & $(0.00001, 63)$ & $(0.01, 0.01, 0.001, 4)$ & $(1, 10, 0.01, 183)$ \\
100000 & $71.67$ & $55$ & $63.33$ & $73.33$ & $78.33$ & $75$ & $61.67$ & $78.33$ \\
 & $(0.0625)$ & $(0.00001, 0.00001, 0.00001)$ & $(100000, 183)$ & $(0.00001, 123)$ & $(0.001, 183)$ & $(0.00001, 123)$ & $(100, 1000, 100, 4)$ & $(0.1, 100, 0.01, 3)$  \\
101000 & $66.67$ & $58.33$ & $75$ & $73.33$ & $68.33$ & $76.67$ & $55$ & $80.00$  \\
 & $(0.0625)$ & $(0.00001, 0.00001, 0.00001)$ & $(0.001, 123)$ & $(0.00001, 143)$ & $(0.001, 3)$ & $(0.00001, 143)$ & $(100, 10, 100, 4)$ & $(1000, 10000, 10, 23)$  \\
102000 & $81.67$ & $51.67$ & $81.67$ & $75$ & $81.67$ & $71.67$ & $83.33$ & $81.67$  \\
 & $(0.03125)$ & $(0.00001, 0.00001, 0.00001)$ & $(0.01, 63)$ & $(0.00001, 143)$ & $(0.01, 3)$ & $(100000, 3)$ & $(0.001, 0.001, 0.0001, 4)$ & $(0.1, 100, 0.00001, 3)$ \\
103000 & $73.33$ & $45$ & $80$ & $75$ & $81.67$ & $73.33$ & $65$ & $80.00$  \\
 & $(0.03125)$ & $(0.00001, 0.00001, 0.00001)$ & $(0.001, 163)$ & $(0.0001, 43)$ & $(0.001, 163)$ & $(0.00001, 43)$ & $(0.0001, 0.1, 0.001, 4)$ & $(1000, 100000, 0.0001, 43)$ \\
104000 & $71.67$ & $46.67$ & $55$ & $65$ & $76.67$ & $76.67$ & $46.67$ & $51.67$  \\
 & $(32)$ & $(0.00001, 0.00001, 0.00001)$ & $(10, 123)$ & $(10000, 23)$ & $(0.0001, 203)$ & $(0.00001, 83)$ & $(0.0001, 0.0001, 0.00001, 4)$ & $(100, 0.1, 1000, 83)$  \\
108000 & $80$ & $48.33$ & $75.65$ & $76.67$ & $78.33$ & $76.67$ & $78.33$ & $81.67$ \\
 & $(8)$ & $(0.00001, 0.00001, 0.00001)$ & $(0.001, 63)$ & $(0.00001, 83)$ & $(0.1, 3)$ & $(0.00001, 3)$ & $(100, 1000, 10000, 4)$ & $(100, 100000, 0.01, 23)$  \\
109000 & $65$ & $43.33$ & $68.33$ & $73.33$ & $68.33$ & $75$ & $73.33$ & $88.33$ \\
 & $(1)$ & $(0.00001, 0.00001, 0.00001)$ & $(0.0001, 203)$ & $(100, 63)$ & $(0.0001, 203)$ & $(0.0001, 43)$ & $(0.001, 0.001, 0.0001, 0.25)$ & $(1, 100, 0.001, 23)$   \\
113000 & $66.67$ & $45$ & $68.33$ & $63.33$ & $70$ & $63.33$ & $66.67$ & $66.67$  \\
 & $(0.03125)$ & $(0.00001, 0.00001, 0.00001)$ & $(0.001, 123)$ & $(1000, 203)$ & $(0.001, 123)$ & $(0.0001, 23)$ & $(0.0001, 0.001, 0.001, 0.25)$ & $(100000, 10000, 0.00001, 3)$  \\
118000 & $58.33$ & $48.33$ & $68.33$ & $61.67$ & $68.33$ & $56.67$ & $58.33$ & $60.00$ \\
 & $(8)$ & $(0.00001, 0.00001, 0.00001)$ & $(0.001, 103)$ & $(100000, 43)$ & $(0.001, 103)$ & $(1000, 43)$ & $(0.01, 0.1, 0.01, 4)$ & $(100000, 100, 1, 183)$  \\
119000 & $71.33$ & $53.33$ & $81.67$ & $83.33$ & $78.33$ & $80$ & $70$ & $80.00$  \\
 & $(16)$ & $(0.00001, 0.00001, 0.00001)$ & $(0.001, 143)$ & $(0.00001, 123)$ & $(0.01, 63)$ & $(0.00001, 83)$ & $(0.001, 0.0001, 0.01, 0.25)$ & $(0.0001, 0.01, 0.00001, 23)$  \\
12000 & $60$ & $48.33$ & $63.33$ & $58.33$ & $63.33$ & $51.67$ & $61.67$ & $68.33$  \\
 & $(32)$ & $(0.00001, 0.00001, 0.00001)$ & $(0.0001, 83)$ & $(0.00001, 63)$ & $(0.0001, 83)$ & $(0.00001, 43)$ & $(0.1, 0.1, 0.1, 2)$ & $(0.00001, 10000, 1000, 203)$  \\
120000 & $73.33$ & $48.33$ & $78.33$ & $80$ & $78.33$ & $85$ & $66.67$ & $75.00$  \\
 & $(0.03125)$ & $(0.00001, 0.00001, 0.00001)$ & $(0.01, 203)$ & $(0.00001, 203)$ & $(0.01, 203)$ & $(0.00001, 103)$ & $(0.001, 0.0001, 0.001, 4)$ & $(0.0001, 10, 100, 203)$  \\
121000 & $63.33$ & $55$ & $68.33$ & $75$ & $71.67$ & $71.67$ & $71.67$ & $75.00$  \\
 & $(2)$ & $(0.00001, 0.00001, 0.00001)$ & $(0.0001, 163)$ & $(0.00001, 103)$ & $(0.001, 143)$ & $(0.00001, 103)$ & $(0.001, 0.1, 0.01, 2)$ & $(0.0001, 0.01, 0.00001, 3)$  \\
122000 & $78.33$ & $55$ & $70$ & $76.67$ & $70$ & $71.67$ & $61.67$ & $81.67$  \\
 & $(0.0625)$ & $(0.00001, 0.00001, 0.00001)$ & $(0.01, 163)$ & $(100, 23)$ & $(0.01, 163)$ & $(0.00001, 123)$ & $(0.01, 0.001, 0.01, 0.25)$ & $(0.0001, 0.1, 0.00001, 23)$  \\
13000 & $85$ & $50$ & $83.33$ & $91.67$ & $85$ & $90$ & $78.33$ & $85$  \\
 & $(32)$ & $(0.00001, 0.00001, 0.00001)$ & $(0.001, 183)$ & $(0.00001, 103)$ & $(0.001, 183)$ & $(0.00001, 43)$ & $(0.01, 0.001, 0.0001, 2)$ & $(0.0001, 0.001, 0.00001, 3)$ \\
130000 & $58.96$ & $48.33$ & $58.33$ & $65$ & $58.33$ & $61.67$ & $60$ & $61.67$  \\
 & $(0.03125)$ & $(0.00001, 0.00001, 0.00001)$ & $(0.001, 103)$ & $(100, 23)$ & $(0.001, 103)$ & $(0.00001, 23)$ & $(0.001, 0.001, 0.001, 4)$ & $(0.001, 0.1, 0.00001, 83)$ \\
131000 & $78.33$ & $63.33$ & $73.33$ & $76.67$ & $83.33$ & $76.67$ & $81.67$ & $75.00$  \\
 & $(0.03125)$ & $(0.00001, 0.00001, 0.00001)$ & $(0.001, 163)$ & $(0.00001, 83)$ & $(0.01, 23)$ & $(0.00001, 183)$ & $(0.01, 0.001, 0.01, 4)$ & $(1, 1000, 1000, 163)$ \\
140000 & $71.67$ & $40$ & $58.33$ & $76.67$ & $56.67$ & $78.33$ & $45$ & $80.00$\\
 & $(0.0625)$ & $(0.00001, 0.00001, 0.00001)$ & $(0.1, 183)$ & $(0.00001, 143)$ & $(0.1, 3)$ & $(100, 3)$ & $(0.01, 0.001, 0.0001, 4)$ & $(10, 1, 1, 3)$  \\
142000 & $86.67$ & $51.67$ & $85$ & $68.33$ & $91.67$ & $88.33$ & $78.33$ & $91.67$  \\
 & $(0.25)$ & $(0.00001, 0.00001, 0.00001)$ & $(0.001, 183)$ & $(10000, 23)$ & $(0.1, 3)$ & $(1, 3)$ & $(0.01, 0.001, 0.0001, 4)$ & $(1, 100, 0.01, 3)$ \\
143000 & $66.67$ & $51.67$ & $63.33$ & $65$ & $63.33$ & $65$ & $71.67$ & $58.33$\\
 & $(0.03125)$ & $(0.00001, 0.00001, 0.00001)$ & $(0.001, 163)$ & $(0.00001, 203)$ & $(0.001, 163)$ & $(0.00001, 203)$ & $(0.01, 0.1, 0.1, 0.25)$ & $(0.001, 10, 1, 183)$  \\
144000 & $73.33$ & $48.33$ & $68.33$ & $71.67$ & $68.33$ & $70$ & $66.67$ & $68.33$ \\
 & $(16)$ & $(0.00001, 0.00001, 0.00001)$ & $(0.01, 63)$ & $(0.00001, 203)$ & $(0.01, 63)$ & $(0.00001, 203)$ & $(0.1, 0.1, 0.1, 0.25)$ & $(100, 0.1, 0.00001, 203)$  \\
147000 & $58.33$ & $55$ & $63.33$ & $73.33$ & $63.33$ & $73.33$ & $70$ & $75.00$  \\
 & $(0.125)$ & $(0.00001, 0.00001, 0.00001)$ & $(0.01, 203)$ & $(0.00001, 123)$ & $(0.01, 203)$ & $(0.00001, 123)$ & $(0.001, 0.001, 0.1, 4)$ & $(0.1, 1, 0.00001, 203)$  \\
148000 & $86.67$ & $50$ & $90$ & $80$ & $88.33$ & $86.67$ & $83.33$ & $83.33$  \\
 & $(2)$ & $(0.00001, 0.00001, 0.00001)$ & $(0.001, 103)$ & $(0.00001, 83)$ & $(0.001, 23)$ & $(0.00001, 63)$ & $(0.001, 0.0001, 0.1, 4)$ & $(0.1, 100, 0.001, 83)$\\
152000 & $56.67$ & $48.33$ & $65$ & $66.67$ & $66.67$ & $45$ & $51.67$ & $68.33$\\
 & $(0.25)$ & $(0.00001, 0.00001, 0.00001)$ & $(0.0001, 143)$ & $(0.1, 23)$ & $(0.00001, 103)$ & $(1, 103)$ & $(0.1, 0.1, 0.001, 0.25)$ & $(100, 100000, 10, 43)$ \\
153000 & $79.33$ & $55$ & $76.67$ & $73.33$ & $76.67$ & $68.33$ & $80.00$ & $71.00$  \\
 & $(0.5)$ & $(0.00001, 0.00001, 0.00001)$ & $(0.001, 203)$ & $(0.00001, 43)$ & $(0.001, 203)$ & $(0.00001, 43)$ & $(0.01, 10, 100, 4)$ & $(1, 1000, 0.1, 203)$  \\
161000 & $86.67$ & $46.67$ & $91.67$ & $91.67$ & $88.33$ & $90$ & $93.33$ & $91.67$ \\
 & $(0.5)$ & $(0.00001, 0.00001, 0.00001)$ & $(0.001, 183)$ & $(0.00001, 143)$ & $(0.1, 23)$ & $(0.00001, 43)$ & $(0.001, 0.0001, 0.00001, 2)$ & $(100, 1000, 0.001, 3)$  \\
163000 & $80$ & $41.67$ & $80$ & $78.33$ & $80$ & $85$ & $71.67$ & $83.33$ \\
 & $(4)$ & $(0.00001, 0.00001, 0.00001)$ & $(0.0001, 203)$ & $(1, 23)$ & $(0.001, 83)$ & $(0.00001, 3)$ & $(0.001, 0.0001, 0.00001, 0.5)$ & $(0.0001, 0.01, 0.00001, 43)$  \\
17000 & $83.33$ & $56.67$ & $86.67$ & $90$ & $86.67$ & $86.67$ & $83.33$ & $93.33$ \\
 & $(2)$ & $(0.00001, 0.00001, 0.00001)$ & $(0.001, 163)$ & $(0.00001, 143)$ & $(0.001, 163)$ & $(0.00001, 203)$ & $(0.001, 0.00001, 0.1, 4)$ & $(0.001, 0.1, 0.0001, 203)$  \\
171000 & $88.33$ & $48.33$ & $76.67$ & $68.33$ & $76.67$ & $75$ & $63.33$ & $75.00$  \\
 & $(0.03125)$ & $(0.00001, 0.00001, 0.00001)$ & $(0.001, 123)$ & $(0.00001, 83)$ & $(0.001, 123)$ & $(0.00001, 3)$ & $(0.1, 0.001, 0.1, 0.25)$ & $(0.00001, 0.01, 0.00001, 183)$  \\
173000 & $85$ & $61.67$ & $86.67$ & $71.67$ & $83.33$ & $78.33$ & $80$ & $86.67$  \\
 & $(0.125)$ & $(0.00001, 0.00001, 0.00001)$ & $(0.001, 203)$ & $(1000, 23)$ & $(0.001, 63)$ & $(1000, 3)$ & $(0.0001, 0.01, 0.1, 0.25)$ & $(10, 10000, 0.00001, 23)$  \\
174000 & $81.67$ & $45$ & $81.67$ & $85$ & $83.33$ & $88.33$ & $88.33$ & $90.00$  \\
 & $(0.03125)$ & $(0.00001, 0.00001, 0.00001)$ & $(0.01, 103)$ & $(0.001, 43)$ & $(0.01, 103)$ & $(0.00001, 23)$ & $(0.0001, 0.01, 0.0001, 4)$ & $(0.00001, 0.0001, 0.00001, 3)$  \\
182000 & $76.67$ & $46.67$ & $78.33$ & $76.67$ & $80$ & $76.67$ & $70$ & $81.67$  \\
 & $(0.03125)$ & $(0.00001, 0.00001, 0.00001)$ & $(0.001, 183)$ & $(0.00001, 123)$ & $(0.001, 183)$ & $(0.00001, 123)$ & $(0.00001, 0.0001, 0.0001, 0.25)$ & $(1, 100, 0.1, 3)$  \\
183000 & $70$ & $48.33$ & $80$ & $68.33$ & $80$ & $78.33$ & $68.33$ & $80.00$  \\
 & $(4)$ & $(0.00001, 0.00001, 0.00001)$ & $(0.0001, 203)$ & $(1000, 43)$ & $(0.0001, 203)$ & $(0.00001, 43)$ & $(0.00001, 0.01, 0.00001, 4)$ & $(1, 10000, 0.0001, 203)$ \\
187000 & $83.33$ & $43.33$ & $81.67$ & $80$ & $83.33$ & $85$ & $81.67$ & $83.33$ \\
 & $(0.03125)$ & $(0.00001, 0.00001, 0.00001)$ & $(0.0001, 103)$ & $(10000, 23)$ & $(0.001, 103)$ & $(0.00001, 3)$ & $(0.01, 0.01, 0.01, 0.25)$ & $(0.0001, 0.1, 0.00001, 163)$  \\
189000 & $76.67$ & $58.33$ & $80$ & $81.67$ & $80$ & $75$ & $73.33$ & $75.00$  \\
 & $(0.25)$ & $(0.00001, 0.00001, 0.00001)$ & $(0.001, 203)$ & $(0.00001, 123)$ & $(0.001, 203)$ & $(0.00001, 163)$ & $(0.01, 0.001, 0.001, 4)$ & $(1000, 0.1, 10, 3)$  \\
20000 & $65$ & $40$ & $63.33$ & $73.33$ & $70$ & $63.33$ & $65$ & $73.33$  \\
 & $(0.03125)$ & $(0.00001, 0.00001, 0.00001)$ & $(0.001, 183)$ & $(100000, 23)$ & $(0.01, 3)$ & $(0.00001, 163)$ & $(0.001, 0.001, 0.0001, 4)$ & $(0.1, 100, 0.00001, 43)$  \\
201000 & $86.67$ & $58.33$ & $90$ & $88.33$ & $91.67$ & $86.67$ & $80$ & $90.00$  \\
 & $(1)$ & $(0.00001, 0.00001, 0.00001)$ & $(1, 23)$ & $(0.00001, 63)$ & $(0.001, 63)$ & $(0.00001, 23)$ & $(0.0001, 0.01, 0.001, 4)$ & $(0.001, 0.01, 0.00001, 23)$  \\
21000 & $90$ & $50$ & $93.33$ & $83.33$ & $88.33$ & $86.67$ & $86.67$ & $88.33$  \\
 & $(1)$ & $(0.00001, 0.00001, 0.00001)$ & $(0.001, 183)$ & $(0.00001, 43)$ & $(0.01, 163)$ & $(0.00001, 103)$ & $(0.1, 0.1, 0.1, 0.25)$ & $(1000, 100000, 0.00001, 203)$  \\
22000 & $70$ & $46.67$ & $70$ & $53.33$ & $73.33$ & $68.33$ & $68.33$ & $73.33$ \\
 & $(0.03125)$ & $(0.00001, 0.00001, 0.00001)$ & $(0.001, 143)$ & $(1, 23)$ & $(0.1, 23)$ & $(0.00001, 43)$ & $(0.001, 0.001, 0.001, 0.25)$ & $(10, 1000, 0.001, 203)$ \\
231000 & $65$ & $55$ & $61.67$ & $60$ & $56.67$ & $60$ & $53.33$ & $53.33$ \\
 & $(0.0625)$ & $(0.00001, 0.00001, 0.00001)$ & $(0.1, 123)$ & $(0.00001, 203)$ & $(0.01, 83)$ & $(0.00001, 203)$ & $(0.1, 0.001, 0.0001, 0.25)$ & $(100, 1000, 0.0001, 123)$  \\
276000 & $78.67$ & $56.67$ & $78.33$ & $76.67$ & $80$ & $76.67$ & $71.67$ & $78.33$ \\
 & $(0.25)$ & $(0.00001, 0.00001, 0.00001)$ & $(0.01, 63)$ & $(0.00001, 183)$ & $(0.01, 63)$ & $(0.00001, 183)$ & $(0.001, 0.0001, 0.0001, 4)$ & $(0.01, 10, 0.00001, 23)$ \\
296000 & $78.33$ & $46.67$ & $85$ & $71.67$ & $85$ & $78.33$ & $68.33$ & $76.67$  \\
 & $(16)$ & $(0.00001, 0.00001, 0.00001)$ & $(0.001, 183)$ & $(0.00001, 103)$ & $(0.001, 183)$ & $(0.00001, 3)$ & $(0.00001, 0.001, 0.01, 4)$ & $(0.01, 10, 0.001, 103)$ \\
33000 & $83.33$ & $43.33$ & $86.67$ & $81.67$ & $88.33$ & $81.67$ & $68.33$ & $88.33$  \\
 & $(0.03125)$ & $(0.00001, 0.00001, 0.00001)$ & $(0.001, 143)$ & $(0.00001, 43)$ & $(0.01, 3)$ & $(0.00001, 43)$ & $(0.01, 0.1, 0.1, 0.25)$ & $(1, 10000, 0.1, 63)$ \\
335000 & $78.33$ & $43.33$ & $75$ & $78.33$ & $73.33$ & $75$ & $70$ & $76.67$  \\
 & $(8)$ & $(0.00001, 0.00001, 0.00001)$ & $(0.001, 203)$ & $(0.00001, 43)$ & $(0.01, 3)$ & $(0.00001, 43)$ & $(0.1, 0.1, 0.1, 4)$ & $(0.00001, 1, 0.00001, 103)$  \\
34000 & $80.33$ & $53.33$ & $76.67$ & $86.67$ & $83.33$ & $86.67$ & $73.33$ & $85.00$ \\
 & $(0.25)$ & $(0.00001, 0.00001, 0.00001)$ & $(0.001, 83)$ & $(0.00001, 163)$ & $(0.001, 203)$ & $(0.00001, 163)$ & $(0.1, 0.001, 0.001, 0.5)$ & $(100000, 10000, 1, 3)$  \\
384000 & $85$ & $55$ & $78.33$ & $86.67$ & $78.33$ & $85$ & $78.33$ & $81.67$  \\
 & $(8)$ & $(0.00001, 0.00001, 0.00001)$ & $(0.001, 83)$ & $(1000, 63)$ & $(0.001, 83)$ & $(0.00001, 183)$ & $(0.01, 0.01, 0.001, 2)$ & $(0.00001, 0.01, 0.00001, 143)$  \\
41000 & $65$ & $43.33$ & $66.67$ & $68.33$ & $56.67$ & $65$ & $56.67$ & $66.67$  \\
 & $(0.125)$ & $(0.00001, 0.00001, 0.00001)$ & $(0.1, 43)$ & $(0.00001, 183)$ & $(0.1, 43)$ & $(1000, 23)$ & $(0.001, 0.0001, 0.01, 4)$ & $(0.00001, 0.01, 0.00001, 143)$ \\
46000 & $70$ & $46.67$ & $78.33$ & $81.67$ & $80$ & $83.33$ & $65$ & $71.67$  \\
 & $(0.125)$ & $(0.00001, 0.00001, 0.00001)$ & $(0.001, 163)$ & $(0.00001, 43)$ & $(0.001, 163)$ & $(0.00001, 43)$ & $(0.0001, 0.001, 0.01, 4)$ & $(1000, 100000, 0.00001, 103)$  \\ \hline
Average ACC & $74.87$ & $49.93$ & $74.98$ & $74.83$ & $76.33$ & $75.43$ & $69.87$ & $77.33$ \\   \hline
Average Rank & $4.11$ & $7.91$ & $3.99$ & $3.97$ & $3.42$ & $4.05$ & $5.61$ & $2.94$  \\ \hline
\end{tabular}}\begin{flushleft}
    \footnotesize{ $^{\dagger}$ represents the proposed model.}
\end{flushleft}
\end{table*}

\renewcommand{\thetable}{S.3}
\begin{table*}[htp]
\centering
    \caption{Performance comparison of the proposed GRVFL-MV along with the baseline models based on classification accuracy for AwA datasets.}
    \label{Classification performance in AwA Case.}
    \resizebox{0.9\linewidth}{!}{
\begin{tabular}{lcccccccc}
\hline
Dataset & SVM2K \cite{farquhar2005two} & MvTSVM \cite{xie2015multi} & {RVFLwoDL1} \cite{huang2006extreme} & {RVFLwoDL2}\cite{huang2006extreme} & RVFL1 \cite{pao1994learning} & RVFL2 \cite{pao1994learning} & MVLDM \cite{hu2024multiview}  & GRVFL-MV$^{\dagger}$  \\ 
 & $(C_1)$ & $(C_1, C_2, D)$ & $(C, N)$ & $(C, N)$ & $(C, N)$ & $(C, N)$ & $(C_1, \nu_1, \nu_2, \sigma)$ & $(C_1, \theta, \rho, N)$\\ \hline 
Chimpanzee vs Giant panda & $84.03$ & $47.22$ & $71.53$ & $72.92$ & $71.53$ & $80.89$ & $72.22$ & $90.97$ \\
 & $(0.00001)$ & $(0.00001, 0.00001, 0.00001)$ & $(0.0001, 123)$ & $(0.00001, 163)$ & $(0.001, 3)$ & $(0.001, 3)$ & $(1000, 0.0001, 0.01, 4)$ & $(0.001, 10, 0.00001, 3)$ \\
Chimpanzee vs Leopard & $80.11$ & $46.53$ & $63.89$ & $83.33$ & $72.92$ & $80.42$ & $68.75$ & $89.58$ \\
 & $(0.00001)$ & $(0.00001, 0.00001, 0.00001)$ & $(1000, 43)$ & $(0.00001, 203)$ & $(0.01, 23)$ & $(0.0001, 3)$ & $(0.001, 0.00001, 0.00001, 4)$ & $(0.001, 100000, 0.00001, 3)$  \\
Chimpanzee vs Persian cat & $70.86$ & $50$ & $79.86$ & $69.44$ & $79.17$ & $80.56$ & $86.11$ & $83.33$\\
 & $(0.0001)$ & $(0.00001, 0.00001, 0.00001)$ & $(0.0001, 183)$ & $(0.00001, 183)$ & $(0.0001, 183)$ & $(0.1, 3)$ & $(100, 0.0001, 0.00001, 4)$ & $(10, 10000, 0.00001, 3)$ \\
Chimpanzee vs Pig & $50.42$ & $51.39$ & $68.75$ & $81.25$ & $69.44$ & $79.17$ & $66.67$ & $83.33$ \\
 & $(0.01)$ & $(0.00001, 0.00001, 0.00001)$ & $(0.00001, 163)$ & $(0.00001, 163)$ & $(0.00001, 163)$ & $(10, 3)$ & $(100000, 0.001, 0.001, 4)$ & $(0.001, 10, 0.0001, 3)$ \\
Chimpanzee vs Hippopotamus & $70.94$ & $54.86$ & $71.53$ & $70.14$ & $72.92$ & $78.47$ & $78.47$ & $78.47$\\
 & $(0.00001)$ & $(0.00001, 0.00001, 0.00001)$ & $(0.00001, 143)$ & $(0.00001, 183)$ & $(0.00001, 143)$ & $(0.001, 3)$ & $(1000, 0.0001, 1, 4)$ & $(0.0001, 10, 0.00001, 3)$\\
Chimpanzee vs Humpback whale & $92.36$ & $81.39$ & $86.81$ & $92.36$ & $88.89$ & $91.14$ & $81.25$ & $96.53$\\
 & $(0.1)$ & $(0.00001, 0.00001, 0.00001)$ & $(0.0001, 83)$ & $(0.00001, 183)$ & $(0.0001, 83)$ & $(0.01, 3)$ & $(10000, 0.001, 0.0001, 4)$ & $(0.001, 100000, 0.001, 23)$ \\
Chimpanzee vs Raccoon & $80.33$ & $63.47$ & $69.44$ & $63.89$ & $73.61$ & $79.86$ & $72.22$ & $83.33$ \\
 & $(0.001)$ & $(0.00001, 0.00001, 100000)$ & $(0.0001, 203)$ & $(0.00001, 123)$ & $(0.001, 3)$ & $(1000, 23)$ & $(10000, 0.00001, 0.01, 4)$ & $(0.001, 1, 0.0001, 3)$  \\
Chimpanzee vs Rat & $77.08$ & $52.78$ & $57.64$ & $75$ & $63.89$ & $71.94$ & $68.06$ & $82.64$ \\
 & $(10)$ & $(0.00001, 0.00001, 0.00001)$ & $(100000, 43)$ & $(0.00001, 183)$ & $(0.001, 163)$ & $(0.001, 3)$ & $(100000, 0.0001, 0.00001, 4)$ &$(0.1, 1000, 0.0001, 43)$ \\
Chimpanzee vs Seal & $70.69$ & $53.47$ & $79.17$ & $76.39$ & $79.17$ & $79.81$ & $75.69$ & $85.42$\\
 & $(0.01)$ & $(0.00001, 0.00001, 0.00001)$ & $(0.0001, 183)$ & $(0.00001, 143)$ & $(0.001, 3)$ & $(0.0001, 3)$ & $(0.001, 0.001, 0.01, 0.25)$ & $(0.001, 100000, 0.01, 3)$ \\
Giant panda vs Leopard & $80.19$ & $54.17$ & $61.11$ & $78.47$ & $72.92$ & $80.11$ & $61.81$ & $90.97$ \\
 & $(0.01)$ & $(0.00001, 0.00001, 0.00001)$ & $(0.0001, 103)$ & $(0.00001, 203)$ & $(0.001, 3)$ & $(0.001, 3)$ & $(100000, 0.00001, 0.01, 0.25)$ & $(0.001, 10, 0.001, 3)$ \\
Giant panda vs Persian cat & $81.81$ & $52.08$ & $77.78$ & $76.39$ & $64.58$ & $80.42$ & $66.67$ & $84.03$ \\
 & $(0.0001)$ & $(0.00001, 0.00001, 0.00001)$ & $(100, 103)$ & $(0.00001, 203)$ & $(0.001, 23)$ & $(100000, 3)$ & $(0.01, 0.00001, 0.01, 4)$ & $(0.1, 1000, 0.01, 23)$\\
Giant panda vs Pig & $80.56$ & $51.39$ & $63.89$ & $75$ & $65.97$ & $79.81$ & $65.97$ & $88.19$ \\
 & $(0.0001)$ & $(0.00001, 0.00001, 0.00001)$ & $(1, 43)$ & $(0.00001, 203)$ & $(0.001, 3)$ & $(0.001, 3)$ & $(1000, 0.00001, 0.01, 0.25)$ & $(0.001, 10, 0.00001, 3)$ \\
Giant panda vs Hippopotamus & $77.78$ & $54.17$ & $74.31$ & $81.25$ & $68.06$ & $71.94$ & $74.31$ &  $84.72$\\
 & $(0.01)$ & $(0.00001, 0.00001, 0.00001)$ & $(0.00001, 183)$ & $(0.00001, 183)$ & $(0.01, 3)$ & $(0.01, 23)$ & $(100000, 0.001, 100, 2)$ & $(0.01, 100, 0.001, 23)$ \\
Giant panda vs Humpback whale & $93.06$ & $46.53$ & $93.06$ & $91.67$ & $93.06$ & $93.22$ & $93.75$ & $81.25$ \\
 & $(0.00001)$ & $(0.00001, 0.00001, 0.00001)$ & $(0.0001, 203)$ & $(0.00001, 203)$ & $(0.0001, 203)$ & $(0.001, 3)$ & $(0.001, 0.00001, 0.01, 4)$ & $(0.001, 1, 0.00001, 3)$ \\
Giant panda vs Raccoon & $80.19$ & $52.78$ & $68.06$ & $74.31$ & $68.75$ & $80.19$ & $64.58$ & $89.58$ \\
 & $(10000)$ & $(0.00001, 0.00001, 0.00001)$ & $(0.0001, 203)$ & $(0.00001, 163)$ & $(0.00001, 203)$ & $(0.001, 3)$ & $(100000, 0.00001, 0.00001, 2)$ & $(0.001, 10, 0.0001, 3)$ \\
Giant panda vs Rat & $83.33$ & $69.31$ & $66.67$ & $76.39$ & $68.06$ & $80.5$ & $70.14$ & $84.72$\\
 & $(1)$ & $(0.00001, 0.00001, 100000)$ & $(10, 103)$ & $(0.00001, 123)$ & $(0.001, 43)$ & $(0.001, 3)$ & $(10000, 0.0001, 0.01, 0.25)$ & $(0.001, 0.1, 0.00001, 43)$  \\
Giant panda vs Seal & $85.89$ & $56.94$ & $80.56$ & $77.08$ & $80.56$ & $80.19$ & $86.81$ & $89.58$ \\
 & $(0.1)$ & $(0.00001, 0.00001, 0.00001)$ & $(0.0001, 203)$ & $(0.00001, 183)$ & $(0.001, 3)$ & $(0.0001, 3)$ & $(100000, 0.001, 0.01, 2)$ & $(1, 1000, 0.0001, 3)$\\
Leopard vs Persian cat & $82.19$ & $79.31$ & $70.83$ & $84.72$ & $77.78$ & $88.19$ & $80.56$ & $90.97$ \\
 & $(0.00001)$ & $(0.00001, 0.00001, 0.00001)$ & $(100, 63)$ & $(0.00001, 203)$ & $(0.001, 23)$ & $(0.001, 3)$ & $(0.00001, 0.00001, 100, 4)$ & $(0.001, 1, 0.001, 3)$ \\
Leopard vs Pig & $75$ & $61.39$ & $61.11$ & $75$ & $66.67$ & $72.17$ & $68.75$ & $78.47$ \\
 & $(0.01)$ & $(0.00001, 0.00001, 0.00001)$ & $(0.0001, 183)$ & $(0.00001, 183)$ & $(0.001, 3)$ & $(0.001, 3)$ & $(0.01, 0.001, 100, 4)$ &  $(0.001, 1, 0.0001, 3)$\\
Leopard vs Hippopotamus & $78.17$ & $50.69$ & $73.61$ & $77.08$ & $74.31$ & $75.94$ & $75$ &  $ 80.56$\\
 & $(10)$ & $(0.00001, 0.00001, 0.00001)$ & $(0.0001, 143)$ & $(0.00001, 143)$ & $(0.0001, 43)$ & $(0.0001, 3)$ & $(10000, 0.0001, 0.001, 4)$ & $(0.01, 10, 0.001, 23)$\\
Leopard vs Humpback whale & $90.75$ & $79.31$ & $89.58$ & $91.67$ & $90.97$ & $90.83$ & $89.58$ & $95.83$  \\
 & $(0.00001)$ & $(0.00001, 0.00001, 0.00001)$ & $(0.00001, 103)$ & $(0.00001, 183)$ & $(0.0001, 143)$ & $(0.001, 23)$ & $(100, 0.00001, 0.01, 4)$ & $(0.001, 10, 0.0001, 3)$  \\
Leopard vs Raccoon & $80.56$ & $55$ & $59.03$ & $57.64$ & $59.03$ & $69.25$ & $56.94$ & $ 79.17$\\
 & $(0.0001)$ & $(0.00001, 0.00001, 0.00001)$ & $(0.0001, 183)$ & $(0.0001, 183)$ & $(0.001, 3)$ & $(0.001, 3)$ & $(0.01, 0.00001, 0.00001, 0.25)$ & $(100, 100000, 0.00001, 23)$  \\
Leopard vs Rat & $76.42$ & $68.61$ & $72.22$ & $76.39$ & $68.06$ & $79.86$ & $65.28$ & $83.33$\\
 & $(0.0001)$ & $(0.00001, 0.00001, 0.00001)$ & $(10000, 43)$ & $(0.00001, 183)$ & $(0.001, 3)$ & $(0.001, 3)$ & $(10000, 0.0001, 0.00001, 0.25)$ & $(0.001, 1, 0.0001, 3)$ \\
Leopard vs Seal & $80.42$ & $63.47$ & $75.69$ & $79.86$ & $75$ & $83.33$ & $81.25$ & $84.72$ \\
 & $(10000)$ & $(0.00001, 0.00001, 0.00001)$ & $(0.0001, 123)$ & $(0.00001, 143)$ & $(0.0001, 123)$ & $(0.001, 43)$ & $(10000, 0.0001, 0.0001, 4)$ & $(10, 100000, 0.01,3)$ \\
Persian cat vs Pig & $70$ & $69.31$ & $63.89$ & $67.36$ & $70.14$ & $74.31$ & $69.44$ & $73.61$ \\
 & $(0.001)$ & $(0.00001, 0.00001, 0.00001)$ & $(0.0001, 183)$ & $(0.00001, 163)$ & $(0.001, 3)$ & $(10000, 3)$ & $(100, 0.00001, 0.01, 4)$ & $(0.01, 1, 0.00001, 3)$ \\
Persian cat vs Hippopotamus & $76.81$ & $76.53$ & $75.69$ & $79.86$ & $77.08$ & $78.94$ & $75.69$ & $80.56$ \\
 & $(0.01)$ & $(0.00001, 0.00001, 0.00001)$ & $(1, 63)$ & $(0.00001, 203)$ & $(0.01, 3)$ & $(1, 3)$ & $(100000, 0.001, 0.00001, 0.25)$ & $(0.01, 10, 0.001, 23)$ \\
Persian cat vs Humpback whale & $71.67$ & $71.39$ & $81.25$ & $88.19$ & $81.94$ & $81.75$ & $85.42$ & $95.14$ \\
 & $(0.00001)$ & $(0.00001, 0.00001, 0.00001)$ & $(0.00001, 143)$ & $(0.00001, 203)$ & $(0.00001, 143)$ & $(1, 23)$ & $(0.01, 0.00001, 0.1, 4)$ & $(0.001, 1, 0.001, 3)$ \\
Persian cat vs Raccoon & $82.64$ & $79.31$ & $73.61$ & $81.25$ & $69.44$ & $71.64$ & $65.97$ & $84.72$ \\
 & $(0.00001)$ & $(0.00001, 0.00001, 0.00001)$ & $(100, 23)$ & $(0.00001, 163)$ & $(0.001, 23)$ & $(0.001, 3)$ & $(100000, 0.00001, 0.001, 2)$ & $(0.001, 0.1, 0.00001, 43)$ \\
Persian cat vs Rat & $60.44$ & $64.17$ & $54.86$ & $56.25$ & $59.72$ & $60.67$ & $56.94$ & $65.97$ \\
 & $(0.001)$ & $(0.00001, 0.00001, 0.00001)$ & $(0.001, 183)$ & $(100000, 103)$ & $(0.001, 3)$ & $(0.001, 3)$ & $(1000, 0.00001, 10, 0.5)$ & $(0.0001, 1, 0.0001, 3)$ \\
Persian cat vs Seal & $80.42$ & $73.47$ & $72.22$ & $71.53$ & $65.97$ & $72.94$ & $83.33$ & $84.72$ \\
 & $(1)$ & $(0.00001, 0.00001, 0.00001)$ & $(0.0001, 183)$ & $(0.00001, 83)$ & $(0.01, 63)$ & $(0.001, 3)$ & $(10000, 0.00001, 0.01, 4)$ & $(0.0001, 10, 0.00001, 3)$ \\
Pig vs Hippopotamus & $71.53$ & $65.83$ & $70.14$ & $65.97$ & $64.58$ & $67.36$ & $72.22$ &  $ 83.33$\\
 & $(0.0001)$ & $(0.00001, 0.00001, 0.00001)$ & $(0.0001, 83)$ & $(0.00001, 203)$ & $(0.0001, 123)$ & $(0.01, 63)$ & $(1000, 0.00001, 0.01, 0.25)$ & $(10, 10000, 0.00001, 3)$ \\
Pig vs Humpback whale & $80.19$ & $77.92$ & $82.64$ & $89.58$ & $82.64$ & $80.19$ & $88.89$ & $82.64$ \\
 & $(0.01)$ & $(0.00001, 0.00001, 0.00001)$ & $(0.00001, 203)$ & $(0.00001, 203)$ & $(0.00001, 203)$ & $(0.001, 3)$ & $(100000, 0.001, 0.0001, 0.25)$ &  $(0.1, 1000, 0.0001, 43)$\\
Pig vs Raccoon & $71.69$ & $69.31$ & $61.11$ & $72.92$ & $64.58$ & $72.22$ & $62.5$ & $81.25$  \\
 & $(0.00001)$ & $(0.00001, 0.00001, 0.00001)$ & $(0.0001, 43)$ & $(0.00001, 183)$ & $(0.0001, 43)$ & $(10000, 3)$ & $(0.0001, 0.00001, 0.0001, 4)$ & $(0.001, 1, 0.00001, 3)$ \\
Pig vs Rat & $71.53$ & $68.61$ & $62.5$ & $59.72$ & $57.64$ & $68.06$ & $64.58$ & $63.89$  \\
 & $(0.01)$ & $(0.00001, 0.00001, 0.00001)$ & $(1000, 63)$ & $(0.0001, 203)$ & $(0.001, 43)$ & $(0.001, 3)$ & $(1000, 0.00001, 0.01, 0.25)$ & $(100, 10000, 0.001, 43)$ \\
Pig vs Seal & $72.69$ & $65.56$ & $70.14$ & $68.06$ & $72.92$ & $74.31$ & $72.92$ & $78.47$ \\
 & $(0.01)$ & $(0.00001, 0.00001, 0.00001)$ & $(0.00001, 163)$ & $(0.00001, 183)$ & $(0.001, 3)$ & $(0.0001, 3)$ & $(0.001, 0.001, 0.00001, 4)$ & $(0.001, 1, 0.00001, 3)$ \\
Hippopotamus vs Humpback whale & $82.03$ & $80.31$ & $77.78$ & $79.86$ & $81.25$ & $82.81$ & $79.86$ & $88.19$ \\
 & $(0.01)$ & $(0.00001, 0.00001, 0.00001)$ & $(0.0001, 143)$ & $(0.00001, 143)$ & $(0.01, 3)$ & $(0.001, 23)$ & $(0.001, 0.0001, 10000, 0.25)$ & $(0.001, 1, 0.00001, 3)$ \\
Hippopotamus vs Raccoon & $78.47$ & $75.14$ & $72.22$ & $70.14$ & $78.47$ & $80.94$ & $75.69$ & $77.78$ \\
 & $(0.01)$ & $(0.00001, 0.00001, 0.00001)$ & $(0.0001, 203)$ & $(0.00001, 183)$ & $(0.001, 3)$ & $(0.001, 23)$ & $(0.00001, 0.001, 1, 4)$ & $(0.01, 1000, 0.001, 43)$  \\
Hippopotamus vs Rat & $75.33$ & $75.83$ & $65.28$ & $71.53$ & $70.14$ & $70.25$ & $64.58$ & $81.25$\\
 & $(100)$ & $(0.00001, 0.00001, 0.00001)$ & $(1000, 43)$ & $(0.00001, 183)$ & $(0.01, 23)$ & $(0.001, 3)$ & $(100000, 0.0001, 0.1, 4)$ & $(0.001, 0.1, 0.001, 3)$  \\
Hippopotamus vs Seal & $69.44$ & $49.31$ & $58.33$ & $65.28$ & $61.81$ & $70.83$ & $60.42$ & $61.81$ \\
 & $(0.01)$ & $(0.00001, 0.00001, 0.00001)$ & $(0.1, 63)$ & $(0.00001, 183)$ & $(0.001, 43)$ & $(0.001, 3)$ & $(0.001, 0.01, 0.00001, 4)$ & $(0.001, 1, 0.00001. 63)$\\
Humpback whale vs Raccoon & $85.67$ & $80.69$ & $84.03$ & $90.97$ & $81.94$ & $82.36$ & $83.33$ & $93.06$ \\
 & $(0.00001)$ & $(0.00001, 0.00001, 10000)$ & $(0.00001, 83)$ & $(0.00001, 123)$ & $(0.0001, 83)$ & $(0.001, 3)$ & $(10000, 0.0001, 0.01, 0.25)$ & $(10, 100000, 0.00001, 3)$ \\
Humpback whale vs Rat & $82.28$ & $80.31$ & $81.94$ & $86.11$ & $81.94$ & $80.58$ & $77.78$ & $90.97$ \\
 & $(0.01)$ & $(0.00001, 0.00001, 0.00001)$ & $(0.0001, 183)$ & $(0.00001, 203)$ & $(0.0001, 183)$ & $(100, 3)$ & $(100000, 0.001, 10, 0.25)$ & $(0.001, 0.01, 0.0001, 3)$ \\
Humpback whale vs Seal & $76.39$ & $72.08$ & $77.78$ & $73.61$ & $78.47$ & $79.17$ & $78.47$ & $80.56$\\
 & $(0.01)$ & $(0.00001, 0.00001, 0.00001)$ & $(0.0001, 143)$ & $(0.00001, 203)$ & $(0.0001, 143)$ & $(0.0001, 3)$ & $(100000, 0.0001, 0.001, 0.25)$ & $(0.01, 100, 0.0001, 3)$ \\
Raccoon vs Rat & $62.22$ & $61.89$ & $59.72$ & $70.83$ & $61.81$ & $60.22$ & $65.28$ &  $70.83$\\
 & $(100)$ & $(0.00001, 0.00001, 0.00001)$ & $(100, 43)$ & $(0.00001, 163)$ & $(0.001, 163)$ & $(0.001, 3)$ & $(100000, 0.00001, 10, 4)$ &  $(0.001, 0.1, 0.001, 3)$ \\
 Raccoon vs Seal & $90.28$ & $75.39$ & $78.47$ & $84.03$ & $82.64$ & $80.28$ & $75.69$ & $88.19$ \\
 & $(100)$ & $(0.00001, 0.00001, 100000)$ & $(0.00001, 183)$ & $(0.00001, 183)$ & $(0.001, 63)$ & $(0.001, 3)$ & $(100000, 0.00001, 0.01, 4)$ & $(100, 100000, 10, 43)$\\
Rat vs Seal & $70.86$ & $65.17$ & $68.75$ & $68.75$ & $68.75$ & $67.83$ & $69.87$ & $81.25$ \\
 & $(0.001)$ & $(0.00001, 0.00001, 0.00001)$ & $(0.00001, 183)$ & $(0.00001, 143)$ & $(0.00001, 183)$ & $(0.01, 3)$ & $(100000, 0.001, 0.01, 0.25)$ & $(0.0001, 0.1, 0.00001, 3)$ \\ \hline
Average ACC & $77.46$ & $64.31$ & $71.74$ & $75.99$ & $72.87$ & $77.46$ & $73.33$ & $83.29$  \\ \hline
Average Rank & $3.46$ & $6.84$ & $5.99$ & $4.32$ & $5.22$ & $3.59$ & $5.02$ & $1.56$  \\ \hline
\end{tabular}}\begin{flushleft}
    \footnotesize{ $^{\dagger}$ represents the proposed model.}
\end{flushleft}
\end{table*}

\end{document}